# Planning with Noisy Probabilistic Relational Rules

**Tobias Lang**                                             TOBIAS.LANG@TU-BERLIN.DE
**Marc Toussaint**                                          MTOUSSAI@CS.TU-BERLIN.DE
*Machine Learning and Robotics Group*
*Technische Universität Berlin*
*Franklinstraße 28/29, 10587 Berlin, Germany*

## Abstract

Noisy probabilistic relational rules are a promising world model representation for several reasons. They are compact and generalize over world instantiations. They are usually interpretable and they can be learned effectively from the action experiences in complex worlds. We investigate reasoning with such rules in grounded relational domains. Our algorithms exploit the compactness of rules for efficient and flexible decision-theoretic planning. As a first approach, we combine these rules with the Upper Confidence Bounds applied to Trees (UCT) algorithm based on look-ahead trees. Our second approach converts these rules into a structured dynamic Bayesian network representation and predicts the effects of action sequences using approximate inference and beliefs over world states. We evaluate the effectiveness of our approaches for planning in a simulated complex 3D robot manipulation scenario with an articulated manipulator and realistic physics and in domains of the probabilistic planning competition. Empirical results show that our methods can solve problems where existing methods fail.

## 1. Introduction

Building systems that act autonomously in complex environments is a central goal of Artificial Intelligence. Nowadays, A.I. systems are on par with particularly intelligent humans in specialized tasks such as playing chess. They are hopelessly inferior to almost all humans, however, in deceivingly simple tasks of everyday-life, such as clearing a desktop, preparing a cup of tea or manipulating chess figures: "The current state of the art in reasoning, planning, learning, perception, locomotion, and manipulation is so far removed from human-level abilities, that we cannot yet contemplate working in an actual domain of interest" (Pasula, Zettlemoyer, & Kaelbling, 2007). Performing common object manipulations is indeed a challenging task in the real world: we can choose from a very large number of distinct actions with uncertain outcomes and the number of possible situations is basically unseizable.

To act in the real world, we have to accomplish two tasks. First, we need to understand how the world works: for example, a pile of plates is more stable if we place the big plates at its bottom; it is a hard job to build a tower from balls; filling tea into a cup may lead to a dirty table cloth. Autonomous agents need to learn such world knowledge from experience to adapt to new environments and not to rely on human hand-crafting. In this paper, we employ a recent solution for learning (Pasula et al., 2007). Once we know about the possible effects of our actions, we face a second challenging problem: how can we use our acquired knowledge in reasonable time to find a sequence of actions suitable to achieve our goals?

This paper investigates novel algorithms to tackle this second task, namely planning. We pursue a model-based approach for planning in complex domains. In contrast to model-free approaches which compute policies directly from experience with respect to fixed goals (also called habit-based decision making), we follow a purposive decision-making approach (Botvinick & An, 2009) and use learned models to plan for the goal and current state at hand. In particular, we simulate the probabilistic effects of action sequences. This approach has interesting parallels in recent neurobiology and cognitive science results suggesting that the behavior of intelligent mammals is driven by internal simulation or emulation: it has been found that motor structures in the cortex are activated during planning, while the execution of motor commands is suppressed (Hesslow, 2002; Grush, 2004).

Probabilistic relational world model representations have received significant attention over the last years. They enable to generalize over object identities to unencountered situations and objects of similar types and to account for indeterministic action effects and noise. We will review several such approaches together with other related work in Section 2. Noisy indeterministic deictic (NID) rules (Pasula et al., 2007) capture the world dynamics in an elegant compact way. They are particularly appealing as they can be learned effectively from experience. The existing approach for planning with these rules relies on growing full look-ahead trees in the grounded domain. Due to the very large action space and the stochasticity of the world, the computational burden to plan just a single action with this method in a given situation can be overwhelmingly large. This paper proposes two novel ways for reasoning efficiently in the grounded domain using learned NID rules, enabling fast planning in complex environments with varying goals. First, we apply the existing *Upper Confidence bounds applied to Trees* (UCT) algorithm (Kocsis & Szepesvari, 2006) with NID rules. In contrast to full-grown look-ahead trees, UCT samples actions selectively, thereby cutting suboptimal parts of the tree early. Second, we introduce the *Probabilistic Relational Action-sampling in DBNs planning Algorithm* (PRADA) which uses probabilistic inference to cope with uncertain action outcomes. Instead of growing look-ahead trees with sampled successor states like the previous approaches, PRADA applies approximate inference techniques to propagate the effects of actions. In particular, we make three contributions with PRADA: *(i)* Following the idea of framing planning as a probabilistic inference problem (Shachter, 1988; Toussaint, Storkey, & Harmeling, 2010), we convert NID rules into a dynamic Bayesian network (DBN) representation. *(ii)* We derive an approximate inference method to cope with the state complexity of a time-slice of the resulting network. Thereby, we can efficiently predict the effects of action sequences. *(iii)* For planning based on sampling action-sequences, we propose a sampling distribution for plans which takes predicted state distributions into account. We evaluate our planning approaches in a simulated complex 3D robot manipulation environment with realistic physics, with an articulated humanoid manipulating objects of different types (see Fig. 4). This domain contains billions of world states and a large number of potential actions. We learn NID rules from experience in this environment and apply them with our planning approaches in different planning scenarios of increasing difficulty. Furthermore, we provide results of our approaches on the planning domains of the most recent international probabilistic planning competition. For this purpose, we discuss the relation between NID rules and the probabilistic planning domain definition language (PPDDL) used for the specification of these domains.

We begin this paper by discussing the related work in Section 2 and reviewing the background of our work, namely stochastic relational representations, NID rules, the formalization of decision-theoretic planning and graphical models in Section 3. In Section 4, we present two planning algorithms that build look-ahead trees to cope with stochastic actions. In Section 5, we introduce PRADA which uses approximate inference for planning. In Section 6, we present our empirical evaluation demonstrating the utility of our planning approaches. Finally, we conclude and outline future directions of research.

## 2. Related Work

The problem of decision-making and planning in stochastic relational domains has been approached in different ways. The field of relational reinforcement learning (RRL) (Džeroski, de Raedt, & Driessens, 2001; van Otterlo, 2009) investigates value functions and Q-functions that are defined over all possible ground states and actions of a relational domain. The key idea is to describe important world features in terms of abstract logical formulas enabling generalization over objects and situations. Model-free RRL approaches learn value functions for states and actions directly from experience. Q-function estimators include relational regression trees (Džeroski et al., 2001) and instance-based regression using distance metrics between relational states such as graph kernels (Driessens, Ramon, & Gärtner, 2006). Model-free approaches enable planning for the specific problem type used in the training examples, e.g. $on(X, Y)$, and thus may be inappropriate in situations where the goals of the agent change quickly, e.g. from $on(X, Y)$ to $inhand(X)$. In contrast, model-based RRL approaches first learn a relational world model from the state transition experiences and then use this model for planning, for example in the form of relational probability trees for individual state attributes (Croonenborghs, Ramon, Blockeel, & Bruynooghe, 2007) or SVMs using graph kernels (Halbritter & Geibel, 2007). The stochastic relational NID rules of Pasula et al. (2007) are a particularly appealing action model representation, as it has been shown empirically that they can learn the dynamics of complex environments.

Once a probabilistic relational world model is available (either learned or handcrafted), one can pursue decision-theoretic planning in different ways. Within the machine learning community, a popular direction of research formalizes the problem as a relational Markov decision process (RMDP) and develops dynamic programming algorithms to compute solutions, i.e. policies over complete state and action spaces. Many algorithms reason in the *lifted* abstract representation without grounding or referring to particular problem instances. Boutilier, Reiter, and Price (2001) introduce Symbolic Dynamic Programming, the first exact solution technique for RMDPs which uses logical regression to construct minimal logical partitions of the state space required to make all necessary value function distinctions. This approach has not been implemented as it is difficult to keep the first-order state formulas consistent and of manageable size. Based on these ideas, Kersting, van Otterlo, and de Raedt (2004) propose an exact value iteration algorithm for RMDPs using logic-programming, called ReBel. They employ a restricted language to represent RMDPs so that they can reason efficiently over state formulas. Hölldobler and Skvortsova (2004) present a first-order value iteration algorithm (FOVIA) using a different restricted language. Karabaev and Skvortsova (2005) extend FOVIA by combining first-order reasoning about actions with a heuristic search restricted to those states that are reachable from the initial state. Wang, Joshi, and Khardon (2008) derive a value iteration algorithm based on using first-order decision diagrams (FODDs) for goal regression. They introduce reduction operators for FODDs to keep the representation small, which may require complex reasoning; an empirical evaluation has not been provided. Joshi, Kersting, and Khardon (2009) apply model checking to reduce FODDs and generalize them to arbitrary quantification.

All these techniques form an interesting research direction as they reason exactly about abstract RMDPs. They employ different methods to ensure exact regression such as theorem proving, logical simplification, or consistency checking. Therefore, principled approximations of these techniques that can discover good policies in more difficult domains are likewise worth investigating. For instance, Gretton and Thiébaux (2004) employ first-order regression to generate a suitable hypothesis language which they then use for policy induction; thereby, their approach avoids formula rewriting and theorem proving, while still requiring model-checking. Sanner and Boutilier (2007, 2009) present a first-order approximate linear programming approach (FOALP). Prior to producing plans, they approximate the value function based on linear combinations of abstract first-order value functions, showing impressive results on solving RMDPs with millions of states. Fern, Yoon, and Givan (2006) consider a variant of approximate policy iteration (API) where they replace the value-function learning step with a learning step in policy space. They make use of a policy-space bias as described by a generic relational knowledge representation and simulate trajectories to improve the learned policy. Kersting and Driessens (2008) describe a non-parametric policy gradient approach which can deal with propositional, continuous and relational domains in a unified way.

Instead of working in the lifted representation, one may reason in the grounded domain. This makes it straightforward to account for two special characteristics of NID rules: the noise outcome and the uniqueness requirement of rules. When grounding an RMDP which specifies rewards only for a set of goal states, one might in principle apply any of the traditional A.I. planning methods used for propositional representations (Weld, 1999; Boutilier, Dean, & Hanks, 1999). Traditionally, planning is often cast as a search problem through a state and action space, restricting oneself to the portion of the state space that is considered to contain goal states and to be reachable from the current state within a limited horizon. Much research within the planning community has focused on deterministic domains and thus can't be applied straightforwardly in stochastic worlds. A common approach for probabilistic planning, however, is to determinize the planning problem and apply deterministic planners (Kuter, Nau, Reisner, & Goldman, 2008). Indeed, FF-Replan (Yoon, Fern, & Givan, 2007) and its extension using hindsight optimization (Yoon, Fern, Givan, & Kambhampati, 2008) have shown impressive performance on many probabilistic planning competition domains. The common variant of FF-Replan considers each probabilistic outcome of an action as a separate deterministic action, ignoring the respective probabilities. It then runs the deterministic Fast-Forward (FF) planner (Hoffmann & Nebel, 2001) on the determinized problem. FF uses a relaxation of the planning problem: it ignores the delete effects of actions and applies clever heuristics to prune the search space. FF-Replan outputs a sequence of actions and expected states. Each time an action execution leads to a state which is not in the plan, FF-Replan has to replan, i.e., recompute a new plan from scratch in the current state. The good performance of FF-Replan in many probabilistic domains has been explained by the structure of these problems (Little & Thiébaux, 2007). It has been argued that FF-Replan should be less appropriate in domains in which the probability of reaching a dead-end is non-negligible and where the outcome probabilities of actions need to be taken into account to construct a good policy.

Many participants of the most recent probabilistic planning competition (IPPC, 2008) extend FF-Replan to deal with the probabilities of action outcomes (see the competition website for brief descriptions of the algorithms). The winner of the competition, RFF (Teichteil-Konigsbuch, Kuter, & Infantes, 2010), computes a robust policy offline by generating successive execution paths leading to the goal using FF. The resulting policy has low probability of failing. LPPFF uses subgoals generated from a determinization of the probabilistic planning problem to divide it into smaller manageable problems. HMDPP's strategy is similar to the all-outcomes-determinization of FF-Replan, but accounts for the probability associated with each outcome. SEH (Wu, Kalyanam, & Givan, 2008) extends a heuristic function of FF-Replan to cope with local optima in plans by using stochastic enforced hill-climbing.

A common approach to reasoning in a more general reward-maximization context which avoids explicitly dealing with uncertainty is to build look-ahead trees by sampling successor states. Two algorithms which follow this idea, namely SST (Kearns, Mansour, & Ng, 2002) and UCT (Kocsis & Szepesvari, 2006), are investigated in this paper.

Another approach by Buffet and Aberdeen (2009) directly optimizes a parameterized policy using gradient descent. They factor the global policy into simple approximate policies for starting each action and sample trajectories to cope with probabilistic effects.

Instead of sampling state transitions, we propose the planning algorithm PRADA in this paper (based on Lang & Toussaint, 2009a) which accounts for uncertainty in a principled way using approximate inference. Domshlak and Hoffmann (2007) propose an interesting planning approach which comes closest to our work. They introduce a probabilistic extension of the FF planner, using complex algorithms for building probabilistic relaxed planning graphs. They construct dynamic Bayesian networks (DBNs) from hand-crafted STRIPS operators and reason about actions and states using weighted model counting. Their DBN representation, however, is inadequate for the type of stochastic relational rules that we use, for the same reasons why the naive DBN model which we will discuss in Sec. 5.1 is inappropriate. Planning by inference approaches (Toussaint & Storkey, 2006) spread information also backwards through DBNs and calculate posteriors over actions (resulting in policies over complete state spaces). How to use backward propagation or even full planning by inference in relational domains is an open issue.

All approaches working in the grounded representation have in common that the number of states and actions will grow exponentially with the number of objects. To apply them in domains with very many objects, these approaches need to be combined with complementary methods that reduce the state and action space complexity in relational domains. For instance, one can focus on envelopes of states which are high-utility subsets of the state space (Gardiol & Kaelbling, 2003), one can ground the representation only with respect to relevant objects (Lang & Toussaint, 2009b), or one can exploit the equivalence of actions (Gardiol & Kaelbling, 2007), which is particularly useful in combination with ignoring certain predicates and functions of the relational logic language (Gardiol & Kaelbling, 2008).

## 3. Background

In this section, we set up the theoretical background for the planning algorithms we will present in subsequent sections. First, we describe relational representations to define world states and actions. Then we will present noisy indeterministic deictic (NID) rules in detail and thereafter define the problem of decision-theoretic planning in stochastic relational domains. Finally, we briefly review dynamic Bayesian networks.

### 3.1 State and Action Representation

A relational domain is represented by a relational logic language $\mathcal{L}$: the set of logical predicates $\mathcal{P}$ and the set of logical functions $\mathcal{F}$ contain the relationships and properties that can hold for domain objects. The set of logical predicates $\mathcal{A}$ comprises the possible actions in the domain. A concrete instantiation of a relational domain is made up of a finite set of objects $\mathcal{O}$. If the arguments of a predicate or function are all concrete, i.e. taken from $\mathcal{O}$, we call it *grounded*. A concrete world state $s$ is fully described as a conjunction of all grounded (potentially negated) predicates and function values. Concrete actions $a$ are described by positive grounded predicates from $\mathcal{A}$. The arguments of predicates and functions can also be abstract logical variables which can represent any object. If a predicate or function has only abstract arguments, we call it *abstract*. Abstract predicates and functions enable generalization over objects and situations. We will speak of *grounding* a formula $\psi$ if we apply a substitution $\sigma$ that maps all of the variables appearing in $\psi$ to objects in $\mathcal{O}$.

A relational model $\mathcal{T}$ of the transition dynamics specifies $P(s'|a,s)$, the probability of a successor state $s'$ if action $a$ is performed in state $s$. In this paper, this is usually a non-deterministic distribution. $\mathcal{T}$ is typically defined compactly in terms of formulas over abstract predicates and functions. This enables abstraction from object identities and concrete domain instantiations. For instance, consider a set of $N$ cups: the effects of trying to grab any of these cups may be described by the same single abstract model instead of using $N$ individual models. To apply $\mathcal{T}$ in a given world state, one needs to ground $\mathcal{T}$ with respect to some of the objects in the domain. NID rules are an elegant way to specify such a model $\mathcal{T}$ and are described in the following.

### 3.2 Noisy Indeterministic Deictic Rules

We want to learn a relational model of a stochastic world and use it for planning. Pasula et al. (2007) have recently introduced an appealing action model representation based on noisy indeterministic deictic (NID) rules which combine several advantages:

- a *relational* representation enabling generalization over objects and situations,

- *indeterministic* action outcomes with probabilities to account for stochastic domains,

- *deictic references* for actions to reduce action space,

- *noise outcomes* to avoid explicit modeling of rare and overly complex outcomes, and

- the existence of an effective *learning algorithm*.

Table 1 shows an exemplary NID rule for our complex robot manipulation domain. Fig. 1 depicts a situation where this rule can be used for prediction. Formally, a NID rule $r$ is given as

$$a_r(\mathcal{X}) : \ \Phi_r(\mathcal{X}) \quad \rightarrow \quad \begin{cases} p_{r,1} & : \quad \Omega_{r,1}(\mathcal{X}) \\ & \vdots \\ p_{r,m_r} & : \quad \Omega_{r,m_r}(\mathcal{X}) \\ p_{r,0} & : \quad \Omega_{r,0} \end{cases} \tag{1}$$

where $\mathcal{X}$ is a set of logical variables in the rule (which represent a (sub-)set of abstract objects). In the rules which define our world models all formulas are abstract, i.e., their arguments are logical variables. The rule $r$ consists of preconditions, namely that action $a_r$ is applied on $\mathcal{X}$ and that the state context $\Phi_r$ is fulfilled, and $m_r + 1$ different outcomes with associated probabilities $p_{r,i} \geq 0$, $\sum_{i=0} p_{r,i} = 1$. Each outcome $\Omega_{r,i}(\mathcal{X})$ describes which predicates and functions change when the rule is applied. The context $\Phi_r(\mathcal{X})$ and outcomes $\Omega_{r,i}(\mathcal{X})$ are conjunctions of (potentially negated) literals constructed from the predicates in $\mathcal{P}$ as well as equality statements comparing functions from $\mathcal{F}$ to constant values. Besides the explicitly stated outcomes $\Omega_{r,i}$ ($i > 0$), the so-called *noise outcome* $\Omega_{r,0}$ models implicitly all other potential outcomes of this rule. In particular, this includes the rare and overly complex outcomes typical for noisy domains, which we do not want to cover explicitly for compactness and generalization reasons. For instance, in the context of the rule depicted in Fig. 1 a potential, but highly improbable outcome is to grab the blue cube while pushing all other objects of the table: the noise outcome allows to account for this without the burden of explicitly stating it.

The arguments of the action $a(\mathcal{X}_a)$ may be a true subset $\mathcal{X}_a \subset \mathcal{X}$ of the variables $\mathcal{X}$ of the rule. The remaining variables are called deictic references $\mathcal{D} = \mathcal{X} \setminus \mathcal{X}_a$ and denote objects relative to the agent or action being performed. Using deictic references has the advantage to decrease the arity of action predicates. This in turn reduces the size of the action space by at least an order of magnitude, which can have significant effects on the planning problem. For instance, consider a binary action predicate which in a world of $n$ objects has $n^2$ groundings in contrast to a unary action predicate which has only $n$ groundings.

As above, let $\sigma$ denote a substitution that maps variables to constant objects, $\sigma : \mathcal{X} \rightarrow \mathcal{O}$. Applying $\sigma$ to an abstract rule $r(\mathcal{X})$ yields a *ground rule* $r(\sigma(\mathcal{X}))$. We say a ground rule $r$ *covers* a state $s$ and a ground action $a$ if $s \models \Phi_r$ and $a = a_r$. Let $\Gamma$ be a set of ground NID rules. We define $\Gamma(a) := \{r \mid r \in \Gamma, a_r = a\}$ to be the set of rules that provide predictions for action $a$. If $r$ is the only rule in $\Gamma(a)$ to cover $a$ and state $s$, we call it the *unique covering rule* for $a$ in $s$. If a state-action pair $(s, a)$ has a unique covering rule $r$, we calculate $P(s' \mid s, a)$ by taking all outcomes of $r$ into account weighted by their respective probabilities,

$$P(s'|s,a) \ = \ P(s'|s,r) \ = \ \sum_{i=1}^{m^r} p_{r,i} \, P(s'|\Omega_{r,i}, s) + p_{r,0} \, P(s'|\Omega_{r,0}, s), \tag{2}$$

where, for $i > 0$, $P(s' \mid \Omega_{r,i}, s)$ is a deterministic distribution that is one for the unique state constructed from $s$ taking the changes of $\Omega_{r,i}$ into account. The distribution given

Table 1: Example NID rule for a complex robot manipulation scenario, which models to try to grab a ball $X$. The cube $Y$ is implicitly defined as the one below $X$ (deictic referencing). $X$ ends up in the robot's hand with high probability, but might also fall on the table. With a small probability something unpredictable happens. Confer Fig. 1 for an example application.

$$
\begin{aligned}
grab(X): \quad & on(X,Y),\ ball(X),\ cube(Y),\ table(Z) \\
\rightarrow \quad & \left\{
\begin{array}{lll}
0.7 & : & inhand(X),\ \neg on(X,Y) \\
0.2 & : & on(X,Z),\ \neg on(X,Y) \\
0.1 & : & \text{noise}
\end{array}
\right.
\end{aligned}
$$

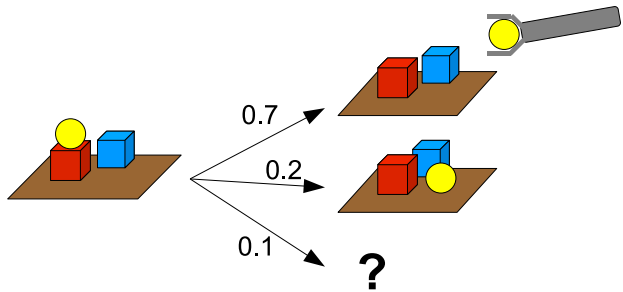

Figure 1: The NID rule defined in Table 1 can be used to predict the effects of action $grab(ball)$ in the situation on the left side. The right side depicts the possible successor states as predicted by the rule. The noise outcome is indicated by a question mark and does not define a unique successor state.

the noise outcome, $P(s'\,|\,\Omega_{r,0},s)$, is unknown and needs to be estimated. Pasula et al. use a worst case constant bound $p_{min} \leq P(s'|\Omega_{r,0},s)$ to lower bound $P(s'|s,a)$. Alternatively, to come up with a well-defined distribution, one may assign very low probability to very many successor states. As described in more detail in Sec. 5.2, our planning algorithm PRADA exploits the factored state representation of a grounded relational domain to achieve this by predicting each state attribute to change with a very low probability.

If a state-action pair $(s,a)$ does *not* have a unique covering rule $r$ (e.g. two rules cover $(s,a)$ providing conflicting predictions), one can predict the effects of $a$ by means of a noisy default rule $r_\nu$ which explains all effects with changing state attributes as noise: $P(s'|s,r_\nu) = P(s'\,|\,\Omega_{r_\nu,0},s)$. Essentially, using $r_\nu$ expresses that we do not know what will happen. This is not meaningful and thus disadvantageous for planning. (Hence, one should bias a NID rules learner to learn rules with contexts which are likely to be mutually exclusive.) For this reason, the concept of unique covering rules is crucial in planning with NID rules. Here, we have to pay the price for using deictic references: when using an abstract NID rule for prediction, we always have to ensure that its deictic references have unique groundings. This may require examining a large part of the state representation, so that proper storage of the ground state and efficient indexing techniques for logical formula evaluation are needed.

The ability to learn models of the environment from experience is a crucial requirement for autonomous agents. The problem of learning rule-sets is in general NP-hard, but efficiency guarantees on the sample complexity can be given for many learning subtasks with suitable restrictions (Walsh, 2010). Pasula et al. (2007) have proposed a supervised batch learning algorithm for *complete* NID rules. This algorithm learns the structure of rules as well as their parameters from experience triples $(s, a, s')$, stating the observed successor state $s'$ after action $a$ was applied in state $s$. It performs a greedy search through the space of rule-sets. It optimizes the tradeoff between maximizing the likelihood of the experience triples and minimizing the complexity of the current hypothesis rule-set $\Gamma$ by optimizing the scoring metric

$$S(\Gamma) = \sum_{(s,a,s')} \log P(s' \mid s, r_{s,a}) - \alpha \sum_{r \in \Gamma} PEN(r) \, , \tag{3}$$

where $r_{s,a}$ is either the unique covering rule for $(s, a)$ or the noisy default rule $r_\nu$ and $\alpha$ is a scaling parameter that controls the influence of regularization. $PEN(r)$ penalizes the complexity of a rule and is defined as the total number of literals in $r$.

The noise outcome of NID rules is crucial for learning. The learning algorithm is initialized with a rule-set comprising only the noisy default rule $r_\nu$ and then iteratively adds new rules or modifies existing ones using a set of search operators. The noise outcome allows avoiding overfitting, as we do not need to model rare and overly complex outcomes explicitly. Its drawback is that its successor state distribution $P(s' \mid \Omega_{r,0}, s)$ is unknown. To deal with this problem, the learning algorithm uses a lower bound $p_{min}$ to approximate this distribution, as described above. This algorithm uses greedy heuristics in its attempt to learn complete rules, so no guarantees on its behavior can be given. Pasula et al., however, report impressive results in complex noisy environments. In Sec. 6.1, we confirm their results in a simulated noisy robot manipulation scenario. Our major motivation for employing NID rules is that we can learn them from observed actions and state transitions. Furthermore, our planning approach PRADA can exploit their simple structure (which is similar to probabilistic STRIPS operators) and convert them into a DBN representation. We provide a detailed comparison of NID rules and PPDDL in Appendix B. While NID rules do not support all features of a sophisticated domain description language such as PPDDL, they can compactly capture the dynamics of many interesting planning domains.

### 3.3 Decision-Theoretic Planning

The problem of decision-theoretic planning is to find actions $a \in \mathcal{A}$ in a given state $s$ which are expected to maximize future rewards for states and actions (Boutilier et al., 1999). In classical planning, this reward is usually defined in terms of a clear-cut goal which is either fulfilled or not fulfilled in a state. This can be expressed by means of a logical formula $\phi$. Typically, this formula is a partial state description so that there exists more than one state where $\phi$ holds. For example, the goal might be to put all our romance books on a specific shelf, no matter where the remaining books are lying. In this case, planning involves finding a sequence of actions $\mathbf{a}$ such that executing $\mathbf{a}$ starting in $s$ will result in a world state $s'$ with $s' \models \phi$. In stochastic domains, however, the outcomes of actions are uncertain. Probabilistic planning is inherently harder than its deterministic counterpart (Littman, Goldsmith, & Mundhenk, 1997). In particular, achieving a goal state with certainty is typically unrealistic. Instead, one may define a lower bound $\theta$ on the probability for achieving a goal state. A second source of uncertainty next to uncertain action outcomes is the uncertainty about the initial state $s$. We will ignore the latter in the following and always assume deterministic initial states. As we will see later, however, it is straightforward to incorporate uncertainty about the initial state using one of our proposed planning approaches.

Instead of a classical planning task which is finished once we have achieved a state where the goal is fulfilled, our task may also be ongoing. For instance, our goal might be to keep the desktop tidy. This can be formalized by means of a reward function over states, which yields high reward for desirable states (for simplicity, here we assume rewards do not depend on actions). This is the approach taken in reinforcement learning formalisms (Sutton & Barto, 1998). Classical planning goals can easily be formalized with such a reward function. We cast the scenario of planning in a stochastic relational domain in a relational Markov decision process (RMDP) framework (Boutilier et al., 2001). We follow the notation of van Otterlo (2009) and define an RMDP as a 4-tuple $(S, A, T, R)$. In contrast to enumerated state spaces, here the state space $S$ has a relational structure defined by logical predicates $\mathcal{P}$ and functions $\mathcal{F}$, which yield the ground atoms with arguments taken from the set of domain objects $\mathcal{O}$. The action space $A$ is defined by positive predicates $\mathcal{A}$ with arguments from $\mathcal{O}$. $T : S \times A \times S \to [0, 1]$ is a transition distribution and $R : S \to \mathbb{R}$ the reward function. Both $T$ and $R$ can make use of the factored relational representation of $S$ and $A$ to abstract from states and actions, as discussed in the following. Typically, the state space $S$ and the action space $A$ of a relational domain are very large. Consider for instance a domain of 5 objects where we use 3 binary predicates to represent states: in this case, the number of states is $2^{3 \cdot 5^2} = 2^{75}$. Relational world models encapsulate the transition probabilities $T$ in a compact way exploiting the relational structure. For example, NID rules as described in Eq. (2) achieve this by generalized partial world state descriptions in the form of conjunctions of abstract literals. The compactness of these models, however, does not carry over directly to the planning problem.

A (deterministic) policy $\pi : S \to A$ tells us which action to take in a given state. For a fixed horizon $d$ and a discount factor $0 < \gamma < 1$, we are interested in maximizing the discounted total reward $r = \sum_{t=0}^{d} \gamma^t r_t$. The value of a factored state is defined as the expected return from state $s$ following policy $\pi$:

$$V^\pi(s) = E[r \mid s_0 = s; \pi] . \tag{4}$$

A solution to an RMDP, and thus to the problem of planning, is an optimal policy $\pi^*$ which maximizes the expected return. It can be defined by the Bellman equation:

$$V^{\pi^*}(s) = R(s) + \gamma \max_{a \in A} [\sum_{s'} P(s' \mid s, a) V^{\pi^*}(s')] . \tag{5}$$

Similarly, one can define the value $Q^\pi(s, a)$ of an action $a$ in state $s$ as the expected return after action $a$ is taken in state $s$, using policy $\pi$ to select all subsequent actions:

$$Q^\pi(s, a) = E[r \,|\, s_0 = s, a_0 = a; \pi] \tag{6}$$

$$= R(s) + \gamma \sum_{s'} V^\pi(s') P(s' \,|\, s, a) \ . \tag{7}$$

The Q-values for the optimal policy $\pi^*$ let us define the optimal action $a^*$ and the optimal value of a state as

$$a^* = \underset{a \in A}{\operatorname{argmax}} \, Q^{\pi^*}(s, a) \quad \text{and} \tag{8}$$

$$V^{\pi^*}(s) = \max_{a \in A} Q^{\pi^*}(s, a) \ . \tag{9}$$

In enumerated unstructured state spaces, state and Q-values can be computed using dynamic programming methods resulting in optimal policies over the complete state space. Recently, promising approaches exploiting relational structure have been proposed that apply similar ideas to solve or approximate solutions in RDMPs on an abstract level (without referring to concrete objects from $\mathcal{O}$) (see related work in Sec. 2). Alternatively, one may reason in the grounded relational domain. This makes it straightforward to account for the noise outcome and the uniqueness requirement of NID rules. Usually, one focuses on estimating the optimal action values for the given state. This approach is appealing for agents with varying goals, where quickly coming up with a plan for the problem at hand is more appropriate than computing an abstract policy over the complete state space. Although grounding simplifies the problem, decision-theoretic planning in the propositionalized representation is a challenging task in complex stochastic domains. In Sections 4 and 5, we present different algorithms reasoning in the grounded relational domain for estimating the optimal Q-values of actions (and action-sequences) for a given state.

### 3.4 Dynamic Bayesian Networks

Dynamic Bayesian networks (DBNs) model the development of stochastic systems over time. The PRADA planning algorithm which we introduce in Sec. 5 makes use of this kind of graphical model to evaluate the stochastic effects of action sequences in factored grounded relational world states. Therefore, we will briefly review Bayesian networks and their dynamic extension here.

A Bayesian network (BN) (Jensen, 1996) is a compact representation of the joint probability distribution over a set of random variables $\mathcal{X}$ by means of a directed acyclic graph $\mathcal{G}$. The nodes in $\mathcal{G}$ represent the random variables, while the edges define their dependencies and thereby express conditional independence assumptions. The value $x$ of a variable $X \in \mathcal{X}$ depends only on the values of its immediate ancestors in $\mathcal{G}$, which are called the parents $Pa(X)$ of X. Conditional probability functions at each node define $P(X \,|\, Pa(X))$. In case of discrete variables, they may be defined in form of conditional probability tables. A BN is a very compact representation of a distribution over $\mathcal{X}$ if all nodes have only few parents or their conditional probability functions have significant local structure. This will play a crucial role in our development of the graphical models for PRADA.

A DBN (Murphy, 2002) extends the BN formalism to model a dynamic system evolving over time. Usually, the focus is on discrete-time stochastic processes. The underlying system itself (in our case, a world state) is represented by a BN $B$, and the DBN maintains a copy of this BN for every time-step. A DBN can be defined as a pair of BNs $(B_0, B_\rightarrow)$, where $B_0$ is a (deterministic or uncertain) prior which defines the state of the system at the initial state $t = 0$, and $B_\rightarrow$ is a two-slice BN which defines the dependencies between two successive time-steps $t$ and $t + 1$. This implements a first-order Markov assumption: the variables at time $t + 1$ depend only on other variables at time $t + 1$ or on variables at $t$.

## 4. Planning with Look-Ahead Trees

To plan with NID rules, one can treat the domain described by the relational logic vocabulary as a relational Markov decision process as discussed in Sec. 3.3. In the following, we present two value-based reinforcement learning algorithms which employ NID rules as a generative model to build look-ahead trees starting from the initial state. These trees are used to estimate the values of actions and states.

### 4.1 Sparse Sampling Trees

The *Sparse Sampling Tree* (SST) algorithm (Kearns et al., 2002) for MDP planning samples randomly sparse, but full-grown look-ahead trees of states starting with the given state as root. This suffices to compute near-optimal actions for any state of an MDP. Given a planning horizon $d$ and a branching factor $b$, SST works as follows (see Fig. 2): In each tree node (representing a state), *(i)* SST takes all possible actions into account, and *(ii)* for each action it takes $b$ samples from the successor state distribution using a generative model for the transitions, e.g. the transition model $T$ of the MDP, to build tree nodes at the next level. Values of the tree nodes are computed recursively from the leaves to the root using the Bellman equation: in a given node, the Q-value of each possible action is estimated by averaging over all values of the $b$ children states for this action; then, the maximizing Q-value over all actions is chosen to estimate the value of the given node. SST has the favorable property that it is independent of the total number of states of the MDP, as it only examines a restricted subset of the state space. Nonetheless, it is exponential in the time horizon taken into account.

Pasula et al. (2007) apply SST for planning with NID rules. When sampling the noise outcome while planning with SST, they assume to stay in the same state, but discount the estimated value. We refer to this adaptation when we speak of SST planning in the remainder of the paper. If an action does not have a unique covering rule, we use the noisy default rule $r_\nu$ to predict its effects. It is always better to perform a *doNothing* action instead where staying in the same state does not get punished. Hence, in SST planning one can discard all actions for a given state which do not have unique covering rules.

While SST is near-optimal, in practice it is only feasible for very small branching factor $b$ and planning horizon $d$. Let the number of actions be $a$. Then the number of nodes at horizon $d$ is $(ba)^d$. (This number can be reduced if the same outcome of a rule is sampled multiple times.) As an illustration, assume we have 10 possible actions per time-step and set parameters $d = 4$ and $b = 4$ (the choice of Pasula et al. in their experiments). To plan a single action for a given state, one has to visit $(10 * 4)^4 = 2,560,000$ states. While smaller choices of $b$ lead to faster planning, they result in a significant accuracy loss in realistic domains. As Kearns et al. note, SST is only useful if no special structure that permits compact representation is available. In Sec. 5, we will introduce an alternative planning approach based on approximate inference that exploits the structure of NID rules.

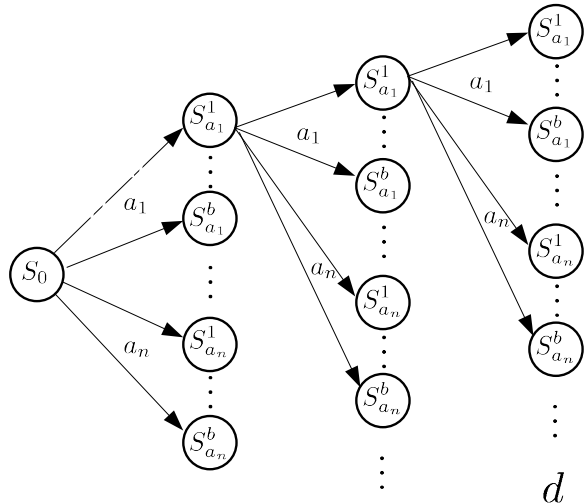

Figure 2: The SST planning algorithm samples sparse, but full-grown look-ahead trees to estimate the values of actions and states.

### 4.2 Sampling Trees with Upper Confidence Bounds

The *Upper Confidence Bounds applied to Trees* (UCT) algorithm (Kocsis & Szepesvari, 2006) also samples a search tree of subsequent states starting with the current state as root. In contrast to SST which generates $b$ successor states for every action in a state, the idea of UCT is to choose actions selectively in a given state and thus to sample selectively from the successor state distribution. UCT tries to identify large subsets of suboptimal actions early in the sampling procedure and to focus on promising parts of the look-ahead tree instead.

UCT builds its look-ahead tree by repeatedly sampling simulated *episodes* from the initial state using a generative model, e.g. the transition model $T$ of the MDP. An episode is a sequence of states, rewards and actions until a limited horizon $d$: $s_0, r_0, a_1, s_1, r_1, a_2 \ldots s_d, r_d$. After each simulated episode, the values of the tree nodes (representing states) are updated online and the simulation policy is improved with respect to the new values. As a result, a distinct value is estimated for each state-action pair in the tree by Monte-Carlo simulation.

More precisely, UCT follows the following policy in tree node $s$: If there exist actions from $s$ which have not been explored yet, then UCT samples one of these using a uniform distribution. Otherwise, if all actions have been explored at least once, then UCT selects the action that maximizes an upper confidence bound $Q^{\triangledown}_{UCT}(s, a)$ on the estimated action value $Q_{UCT}(s, a)$,

$$Q_{UCT}^{\triangledown}(s, a) \;=\; Q_{UCT}(s, a) + c \sqrt{\frac{\log n_s}{n_{s,a}}} \;, \tag{10}$$

$$\pi_{UCT}(s) \;=\; \underset{a}{\operatorname{argmax}} \, Q_{UCT}^{\triangledown}(s, a) \;, \tag{11}$$

where $n_{s,a}$ counts the number of times that action $a$ has been selected from state $s$, and $n_s$ counts the total number of visits to state $s$, $n_s = \sum_a n_{s,a}$. The bias parameter $c$ defines the influence of the number of previous action selections and thereby controls the extent of the upper confidence bound.

At the end of an episode, the value of each encountered state-action pair $(s_t, a_t)$, $0 \leq t < d$, is updated using the total discounted rewards:

$$n_{s_t, a_t} \;\leftarrow\; n_{s_t, a_t} + 1 \;, \tag{12}$$

$$Q_{UCT}(s_t, a_t) \;\leftarrow\; Q_{UCT}(s_t, a_t) + \frac{1}{n_{s_t, a_t}} [\sum_{t'=t}^{d} \gamma^{t'-t} r_{t'} - Q_{UCT}(s_t, a_t)] \;. \tag{13}$$

The policy of UCT implements an exploration-exploitation tradeoff: It balances between exploring currently suboptimal-looking actions that have been selected seldom thus far and exploiting currently best-looking actions to get more precise estimates of their values. The total number of episodes controls the accuracy of UCT's estimates and has to be balanced with its overall running time.

UCT has achieved remarkable results in challenging domains such as the game of Go (Gelly & Silver, 2007). To the best of our knowledge, we are the first to apply UCT for planning in stochastic relational domains, using NID rules as a generative model. We adapt UCT to cope with noise outcomes in the same fashion as SST: we assume to stay in the same state and discount the obtained rewards. Thus, UCT takes only actions with unique covering rules into account, for the same reasons as SST does.

## 5. Planning with Approximate Inference

Uncertain action outcomes characterize complex environments, but make planning in relational domains substantially more difficult. The sampling-based approaches discussed in the previous section tackle this problem by repeatedly generating samples from the outcome distribution of an action using the transition probabilities of an MDP. This leads to look-ahead trees that easily blow up with the planning horizon. Instead of sampling successor states, one may maintain a distribution over states, a so-called "belief". In the following, we introduce an approach for planning in grounded stochastic relation domains which propagates beliefs over states in the sense of state monitoring. First, we show how to create compact graphical models for NID rules. Then we develop an approximate inference method to efficiently propagate beliefs. With this in hand, we describe our *Probabilistic Relational Action-sampling in DBNs planning Algorithm* (PRADA), which samples action-sequences in an informed way and evaluates these using approximate inference in DBNs. Then, an example is presented to illustrate the reasoning of PRADA. Finally, we discuss PRADA in comparison to the approaches of the previous section, SST and UCT, and present a simple extension of PRADA.

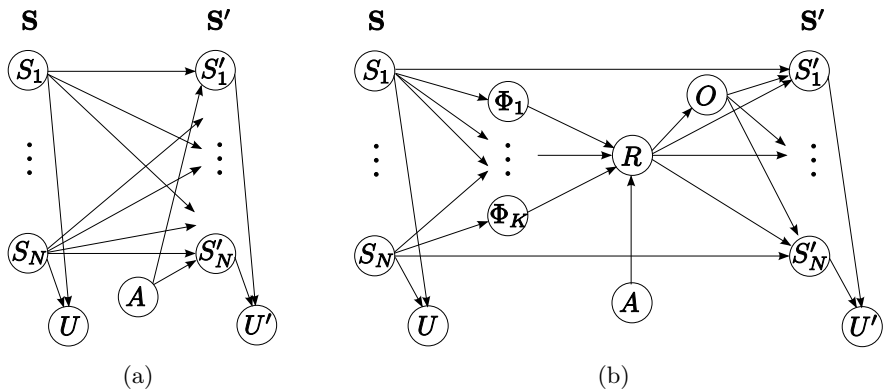

Figure 3: Graphical models for NID rules: (a) Naive DBN; (b) DBN exploiting NID factorization

## 5.1 Graphical Models for NID Rules

Decision-theoretic problems where agents need to choose appropriate actions can be represented by means of Markov chains and dynamic Bayesian networks (DBNs) which are augmented by decision nodes to specify the agent's actions (Boutilier et al., 1999). In the following, we discuss how to convert NID rules to DBNs which the PRADA algorithm will use to plan with probabilistic inference. We denote random variables by upper case letters (e.g. $S$), their values by the corresponding lower case letters (e.g., $s \in dom(S)$), variable vectors by bold upper case letters (e.g. $\mathbf{S} = (S_1, S_2, S_3)$) and value vectors by bold lower case letters (e.g. $\mathbf{s} = (s_1, s_2, s_3)$). We also use column notation, e.g. $\mathbf{s}^{2:4} = (s_2, s_3, s_4)$.

A naive way to convert NID rules to DBNs is shown in Fig. 3(a). States are represented by a vector $\mathbf{S} = (S_1, \ldots, S_N)$ where for each ground predicate in $\mathcal{P}$ there is a binary $S_i$ and for each ground function in $\mathcal{F}$ there is an $S_j$ with range according to the represented function. Actions are represented by an integer variable $A$ which indicates the action out of a vector of ground action predicates in $\mathcal{A}$. The reward gained in a state is represented by $U$ and may depend only on a subset of the state variables. It is possible to express arbitrary reward expectations $P(U \,|\, \mathbf{S})$ with binary $U$ (Cooper, 1988). How can we define the transition dynamics using NID rules in this naive model? Assume we are given a set of fully abstract NID rules. We compute all groundings of these rules w.r.t. the objects of the domain and get the set $\Gamma$ of $K$ different ground NID rules. The parents of a state variable $S_i'$ at the successor time-step include the action variable $A$ and the respective variable $S_i$ at the predecessor time-step. The other parents of $S_i'$ are determined as follows: For each rule $r \in \Gamma$ where the literal corresponding to $S_i'$ appears in the outcomes of $r$, all variables $S_k$ corresponding to literals in the preconditions of $r$ are parents of $S_i'$. As typically $S_i'$ can be manipulated by several actions which in turn are modeled by several rules, the total number of parents of $S_i'$ can be very large. This problem is worsened by the usage of deictic references in the NID rules, as they increase the total number $K$ of ground rules in $\Gamma$. The resulting local structure of the conditional probability function of $S_i'$ is very complex, as one has to account for the uniqueness of covering rules. These complex dependencies between two time-slices make this representation unfeasible for planning.

Therefore, we exploit the structure of NID rules to model a state transition with the compact graphical model shown in Fig. 3(b) representing the joint distribution

$$P(u', \mathbf{s}', o, r, \boldsymbol{\phi} \,|\, a, \mathbf{s}) \quad = \quad P(u' \,|\, \mathbf{s}') \; P(\mathbf{s}' \,|\, o, r, \mathbf{s}) \; P(o \,|\, r) \; P(r \,|\, a, \boldsymbol{\phi}) \; P(\boldsymbol{\phi} \,|\, \mathbf{s}) \,, \qquad (14)$$

which we will explain in detail in the following. As before, assume we are given a set of fully abstract NID rules, for which we compute the set $\Gamma$ of $K$ different ground NID rules w.r.t. the objects in the domain. In addition to $\mathbf{S}$, $\mathbf{S}'$, $A$, $U$ and $U'$ as above, we use a binary random variable $\Phi_i$ for each rule to model the event that its context holds, which is the case if all required literals hold. Let $I(\cdot)$ be the indicator function which is 1 if the argument evaluates to true and 0 otherwise. Then, we have

$$P(\boldsymbol{\phi} \,|\, \mathbf{s}) = \prod_{i=1}^{K} P(\phi_i | \mathbf{s}_{\pi(\Phi_i)}) = \prod_{i=1}^{K} I\left( \bigwedge_{j \in \pi(\Phi_i)} S_j = s_{r_i,j} \right). \qquad (15)$$

We use $\bigwedge_i \rho_i$ to express a logical conjunction $\rho_1 \wedge \cdots \wedge \rho_n$. The function $\pi(\Phi)$ yields the set of indices of the state variables in $\mathbf{s}$, on which $\Phi$ depends. $\mathbf{s}_{r_i}$ denotes the configuration of the state variables corresponding to the literals in the context of $r_i$. We use an integer-valued variable $R$ ranging over $K+1$ possible values to identify the rule which predicts the effects of the action. If it exists, this is the unique covering rule for the current state-action pair, i.e., the only rule $r \in \Gamma(a)$ modeling action $a$ whose context holds:

$$P(R=r|a, \boldsymbol{\phi}) = I\left( r \in \Gamma(a) \wedge \Phi_r = 1 \wedge \bigwedge_{r' \in \Gamma(a) \setminus \{r\}} \Phi_{r'} = 0 \right). \qquad (16)$$

If no unique covering rule exists, we predict no changes as indicated by the special value $R=0$ (assuming not to execute the action, similarly as SST and UCT do):

$$P(R=0 \,|\, a, \boldsymbol{\phi}) = \bigwedge_{r \in \Gamma(a)} \neg I\left( \Phi_r = 1 \wedge \bigwedge_{r' \in \Gamma(a) \setminus \{r\}} \Phi_{r'} = 0 \right). \qquad (17)$$

The integer-valued variable $O$ represents the outcome of the action as predicted by the rule. It ranges over $M$ possible values where $M$ is the maximum number of outcomes all rules in $\Gamma$ have. To ensure a sound semantics, we introduce empty dummy outcomes with zero-probability for those rules whose number of outcomes is less than $M$. The probability of an outcome is defined as in the corresponding rule:

$$P(O=o \,|\, r) = p_{r,o} \,. \qquad (18)$$

We define the probability of the successor state as

$$P(\mathbf{s}' \,|\, o, \mathbf{s}, r) = \prod_i P(s_i' \,|\, o, s_i, r) \,, \qquad (19)$$

which is one for the unique state that is constructed from $\mathbf{s}$ taking the changes according to $\Omega_{r,o}$ into account: if outcome $o$ specifies a value for $S_i'$, this value will have probability one. Otherwise, the value of this state variable persists from the previous time-step. As rules usually change only a small subset of $\mathbf{s}$, persistence most often applies. The resulting dependency $P(s'_i \,|\, o, r, s_i)$ of a variable $S'_i$ at time-step $t + 1$ is compact. In contrast to the naive DBN in Fig. 3(a), it has only three parents, namely the variables for the outcome, the rule and its predecessor at the previous time-step. This simplifies the specification of a conditional probability function for $S'$ significantly and enables efficient inference, as we will see later. The probability of the reward is given by

$$P(U' \!=\! 1 \,|\, \mathbf{s}') = I \left( \bigwedge_{j \in \pi(U')} S'_j \!=\! \tau_j \right) . \tag{20}$$

The function $\pi(U')$ yields the set of indices of the state variables in $\mathbf{s}'$, on which $U'$ depends. The configuration of these variables that corresponds to our planning goal is denoted by $\boldsymbol{\tau}$. Uncertain initial states can be naturally accounted for by specifying priors $P(\mathbf{s}^0)$. We renounce the specification of a prior here, however, as the initial state $\mathbf{s}^0$ will always be given in our experiments later to enable comparison to the look-ahead tree based approaches SST and UCT which require deterministic initial states (which might also be sampled from a prior). Our choice for the distribution $P(a)$ used for sampling actions will be described in Sec. 5.3.

For simplicity we have ignored derived predicates and functions which are defined in terms of other predicates or functions in the presentation of our graphical model. Derived concepts may increase the compactness of rules. If dependencies among concepts are acyclic, it is straightforward to include derived concepts in our model by intra-state dependencies for the corresponding variables. Indeed, we will use derived predicates in our experiments.

We are interested in inferring posterior state distributions $P(\mathbf{s}^t \,|\, \mathbf{a}^{0:t-1})$ given the sequence of previous actions (where we omit conditioning on the initial state for simplicity). Exact inference is intractable in our graphical model. When constructing a junction tree, we will get cliques that comprise whole Markov slices (all variables representing the state at a certain time-step): consider eliminating all state variables $\mathbf{S}^{t+1}$. Due to moralization, the outcome variable $O$ will be connected to all state variables in $\mathbf{S}^t$. After elimination of $O$, all variables in $\mathbf{S}^t$ will form a clique. Thus, we have to make use of approximate inference techniques. General loopy belief propagation (LBP) is unfeasible due to the deterministic dependencies in small cycles which inhibit convergence. We also conducted some preliminary tests in small networks with a damping factor, but without success. It is an interesting open question whether there are ways to alternate between propagating deterministic information and running LBP on the remaining parts of the network, e.g., whether methods such as MC-SAT (Poon & Domingos, 2007) can be successfully applied in decision-making contexts as ours. In the next subsection, we propose a different approximate inference scheme using a factored frontier (FF). The FF algorithm describes a forward inference procedure that computes exact marginals in the next time-step subject to a factored approximation of the previous time-step. Here, our advantage is that we can exploit the structure of the involved DBNs to come up with formulas for these marginals. FF is related to passing only forward messages. In contrast to LBP, information is not propagated backwards. Note that our approach does not condition on rewards (as in full planning by inference) and samples actions, so that backward reasoning is uninformative.

## 5.2 Approximate Inference

In the following, we present an efficient method for approximate inference in the previously proposed DBNs exploiting the factorization of NID rules. We focus on the mathematical derivations. An illustrative example will be provided in Sec. 5.4.

We follow the idea of the factored frontier (FF) algorithm (Murphy & Weiss, 2001) and approximate the belief with a product of marginals:

$$P(\mathbf{s}^t \,|\, \mathbf{a}^{0:t-1}) \quad \approx \quad \prod_i P(s_i^t \,|\, \mathbf{a}^{0:t-1}) \,. \tag{21}$$

We define

$$\alpha(s_i^t) \quad := \quad P(s_i^t \,|\, \mathbf{a}^{0:t-1}) \quad \text{and} \tag{22}$$

$$\alpha(\mathbf{s}^t) \quad := \quad P(\mathbf{s}^t \,|\, \mathbf{a}^{0:t-1}) \approx \prod_{i=1}^N \alpha(s_i^t) \tag{23}$$

and derive a FF filter for the DBN model in Fig. 3(b). We are interested in inferring the state distribution at time $t+1$ given an action sequence $\mathbf{a}^{0:t}$ and calculate the marginals of the state attributes as

$$\alpha(s_i^{t+1}) = P(s_i^{t+1} \,|\, \mathbf{a}^{0:t}) \tag{24}$$

$$= \sum_{r^t} P(s_i^{t+1} \,|\, r^t, \mathbf{a}^{0:t-1}) \, P(r^t \,|\, \mathbf{a}^{0:t}) \,. \tag{25}$$

In Eq. (25), we use all rules for prediction, weighted by their respective posteriors $P(r^t \,|\, \mathbf{a}^{0:t})$. This reflects the fact that depending on the state we use different rules to model the same action. The weight $P(r^t \,|\, \mathbf{a}^{0:t})$ is 0 for all rules not modeling action $a^t$. For the remaining rules which do model $a^t$, the weights correspond to the posterior over those parts of the state space where the according rule is used for prediction.

We compute the first term in (25) as

$$P(s_i^{t+1} \,|\, r^t, \mathbf{a}^{0:t-1}) = \sum_{s_i^t} P(s_i^{t+1} \,|\, r^t, s_i^t) \, P(s_i^t \,|\, r^t, \mathbf{a}^{0:t-1})$$

$$\approx \sum_{s_i^t} P(s_i^{t+1} \,|\, r^t, s_i^t) \, \alpha(s_i^t) \,. \tag{26}$$

Here, we sum over all possible values of the variable $S_i$ at the previous time-step $t$. Intuitively, we take into account all potential "pasts" to arrive at value $s_i^{t+1}$ at the next time-step. The resulting term $P(s_i^{t+1} \,|\, r^t, s_i^t)$ enables us to easily predict the probabilities at the next time-step as discussed below. Each such prediction is weighted by the marginal $\alpha(s_i^t)$ of the respective previous value. The approximation in (26) assumes that $s_i^t$ is conditionally independent of $r^t$. This is not true in general as the choice of a rule for prediction depends on the current state and thus also on attribute $S_i$. To improve on this approximation one can examine whether $s_i^t$ is part of the context of $r^t$: if this is the case, we can infer the state of $s_i^t$ from knowing $r^t$. However, we found our approximation to be sufficient.

As one would expect, we calculate the successor state distribution $P(s_i^{t+1} \,|\, r^t, s_i^t)$ by taking the different outcomes $o$ of $r^t$ into account weighted by their respective probabilities $P(o \,|\, r^t)$,

$$P(s_i^{t+1} \,|\, r^t, s_i^t) = \sum_o P(s_i^{t+1} \,|\, o, r^t, s_i^t) \; P(o \,|\, r^t) \;. \tag{27}$$

This shows us how to update the belief over $S_i^{t+1}$ if we predict with rule $r^t$. $P(s_i^{t+1} \,|\, o, r^t, s_i^t)$ is a deterministic distribution. If $o$ changes the value of $S_i$, $s_i^{t+1}$ is set accordingly. Otherwise, the value $s_i^t$ persists.

Let's turn to the computation of the second term in Eq. (25), $P(r^t \,|\, \mathbf{a}^{0:t})$, the posterior over rules. The trick is to use the context variables $\boldsymbol{\Phi}$ and to exploit the assumption that a rule $r$ models the state transition if and only if it uniquely covers $(a^t, \mathbf{s}^t)$, which is indicated by an appropriate assignment of the $\boldsymbol{\Phi}$. This can then be further reduced to an expression involving only the marginals $\alpha(\cdot)$. We start with

$$
\begin{aligned}
P(R^t = r \,|\, \mathbf{a}^{0:t}) &= \sum_{\boldsymbol{\phi}^t} P(R^t = r \,|\, \boldsymbol{\phi}^t, \mathbf{a}^{0:t}) \; P(\boldsymbol{\phi}^t \,|\, \mathbf{a}^{0:t}) \\
&= I(r \in \Gamma(a^t)) \; P\left(\Phi_r^t = 1, \bigwedge_{r' \in \Gamma(a^t) \setminus \{r\}} \Phi_{r'}^t = 0 \,|\, \mathbf{a}^{0:t-1}\right) \\
&= I(r \in \Gamma(a^t)) \; P(\Phi_r^t = 1 \,|\, \mathbf{a}^{0:t-1}) \; P\left(\bigwedge_{r' \in \Gamma(a^t) \setminus \{r\}} \Phi_{r'}^t = 0 \,|\, \Phi_r^t = 1, \mathbf{a}^{0:t-1}\right) \;.
\end{aligned}
\tag{28}
$$

To simplify the summation over $\boldsymbol{\phi}^t$, we only have to consider the unique assignment of the context variables when $r$ is used for prediction: provided it models the action, as indicated by $I(r \in \Gamma(a^t))$, this is the case if its context $\Phi_r^t$ holds, while the contexts $\Phi_{r'}^t$ of all other "competing" rules $r'$ for action $a^t$ do not hold.

We calculate the second term in (28) by summing over all states $\mathbf{s}$ as

$$P(\Phi_r^t = 1 \,|\, \mathbf{a}^{0:t-1}) = \sum_{\mathbf{s}^t} P(\Phi_r^t = 1 \,|\, \mathbf{s}^t) \; \alpha(\mathbf{s}^t) \approx \sum_{\mathbf{s}^t} P(\Phi_r^t = 1 \,|\, \mathbf{s}^t) \prod_j \alpha(s_j^t) \tag{29}$$

$$= \prod_{j \in \pi(\Phi_r^t)} \alpha(S_j^t = s_{r,j}) \quad. \tag{30}$$

The approximation in (29) is the FF assumption. In (30), $\mathbf{s}_r$ denotes the configuration of the state variables according to the context of $r$ like in (15). We sum out all variables not in the context of $r$. Only the variables in $r$'s context remain: the terms $\alpha(S_j^t = s_{r,j})$ correspond to the probabilities of the respective literals.

The third term in (28) is the joint posterior over the contexts of the competing rules $r'$ given that $r$'s context already holds. We are interested in the situation where none of these other contexts hold. We calculate this as

$$P\left(\bigwedge_{r' \in \Gamma(a^t) \setminus \{r\}} \Phi_{r'}^t = 0 \,|\, \Phi_r^t = 1, \mathbf{a}^{0:t-1}\right) \approx \prod_{r' \in \Gamma(a^t) \setminus \{r\}} P(\Phi_{r'}^t = 0 \,|\, \Phi_r^t = 1, \mathbf{a}^{0:t-1}) \quad, \tag{31}$$

approximating it by the product of the individual posteriors. The latter are computed as

$$P(\Phi_{r'}^t = 0 \mid \Phi_r^t = 1, \mathbf{a}^{0:t-1}) = \sum_{\mathbf{s}^t} P(\Phi_{r'}^t = 0 \mid \mathbf{s}^t) \ P(\mathbf{s}^t \mid \Phi_r^t = 1, \mathbf{a}^{0:t-1}) \tag{32}$$

$$\approx \begin{cases} 1.0 & \text{if} \quad \Phi_r \wedge \Phi_{r'} \to \perp \\ 1.0 - \prod_{\substack{i \in \pi(\Phi_{r'}^t), \\ i \notin \pi(\Phi_r^t)}} \alpha(S_i^t = s_{r',i}) & \text{otherwise} \end{cases}, \tag{33}$$

where the if-condition expresses a logical contradiction of the contexts of $r$ and $r'$. If their contexts contradict, then $r'$'s context will surely not hold given that $r$'s context holds. Otherwise, we know that the state attributes appearing in the contexts of both $r$ and $r'$ do hold as we condition on $\Phi_r = 1$. Therefore, we only have to examine the remaining state attributes of $r'$'s context. Again, we approximate this posterior with the FF marginals.

Finally, we compute the reward probability straightforwardly as

$$P(U^t = 1 \mid \mathbf{a}^{0:t-1}) = \sum_{\mathbf{s}_t} P(U^t = 1 \mid \mathbf{s}_t) P(\mathbf{s}^t \mid \mathbf{a}^{0:t-1}, \mathbf{s}^0) \approx \prod_{i \in \pi(U^t)} \alpha(S_i^t = \tau_i), \tag{34}$$

where $\boldsymbol{\tau}$ denotes the configuration of state variables corresponding to the planning goal as in (20). As above, the summation over states is simplified by the FF assumption resulting in a product of the marginals of the required state attributes.

The overall computational costs of propagating the effects of an action are quadratic in the number of rules for this action (for each such rule we have to calculate the probability that none of the others applies) and linear in the maximum numbers of context literals and manipulated state attributes of those rules.

Our inference framework requires an approximation for the distribution $P(s' \mid \Omega_{r,0}, s)$ (cf. Eq. (2)) to cope with the noise outcome of NID rules. From the training data used to learn rules, we estimate which predicates and functions change value over time as follows: let $\mathbf{S}_c \subset \mathbf{S}$ contain the corresponding variables. We estimate for each rule $r$ the average number $N^r$ of changed state attributes when the noise outcome applies. Due to our factored frontier approach, we can consider the noise effects for each variable independently. We approximate the probability that $S_i \in \mathbf{S}_c$ changes in $r$'s noise outcome by $\frac{N^r}{|S^C|}$. In case of change, all changed values of $S_i$ have equal probability.

## 5.3 Planning

The DBN representation in Fig. 3(b) together with the approximate inference method described in the last subsection enable us to derive a novel planning algorithm for stochastic relational domains: The *Probabilistic Relational Action-sampling in DBNs planning Algorithm* (PRADA) plans by sampling action sequences in an informed way based on predicted beliefs over states and evaluating these action sequences using approximate inference.

More precisely, we sample sequences of actions $\mathbf{a}^{0:T-1}$ of length $T$. For $0 < t \leq T$, we infer the posteriors over states $P(\mathbf{s}^t \mid \mathbf{a}^{0:t-1}, \mathbf{s}^0)$ and rewards $P(u^t \mid \mathbf{a}^{0:t-1}, \mathbf{s}^0)$ (in the sense of filtering or state monitoring). Then, we calculate the value of an action sequence with a discount factor $0 < \gamma < 1$ as

$$Q(\mathbf{s}^0, \mathbf{a}^{0:T-1}) := \sum_{t=0}^{T} \gamma^t P(U^t = 1 \mid \mathbf{a}^{0:t-1}, \mathbf{s}^0) . \tag{35}$$

We choose the first action of the best sequence $\mathbf{a}^* = \text{argmax}_{\mathbf{a}^{0:T-1}} Q(\mathbf{a}^{0:T-1}, s^0)$, if its value exceeds a certain threshold $\theta$ (e.g., $\theta = 0$). Otherwise, we continue sampling action-sequences until either an action is found or planning is given up. The quality of the found plan can be controlled by the total number of action-sequence samples and has to be traded off with the time that is available for planning.

We aim for a strategy to sample good action sequences with high probability. We propose to choose with equal probability among the actions that have a unique covering rule for the current state. Thereby, we avoid the use of the noisy default rule $r_\nu$ which models action effects as noise and is thus of poor use in planning. For the action at time $t$, PRADA samples from the distribution

$$P_{sample}^t(a) \propto \sum_{r \in \Gamma(a)} P\left(\phi_r^t = 1, \bigwedge_{r' \in \Gamma(a) \setminus \{r\}} \phi_{r'}^t = 0 \,|\, \mathbf{a}^{0:t-1}\right). \tag{36}$$

This is a sum over all rules for action $a$: for each such rule we add the posterior that it is the unique covering rule, i.e. that its context $\phi_r^t$ holds, while the contexts $\phi_{r'}^t$ of the competing rules $r'$ do not hold. This sampling distribution takes the current state distribution into account. Thus, the probability to sample an action sequence $\mathbf{a}$ predicting the state sequence $s_0, \ldots, s_T$ depends on the likelihood of the state sequence given $\mathbf{a}$: the more likely the required outcomes are, the more likely the next actions will be sampled. Using this policy, PRADA does not miss actions which SST and UCT explore, as the following proposition states (proof in Appendix A).

**Proposition 1:** *The set of action sequences PRADA samples with non-zero probability is a super-set of the ones of SST and UCT.*

In our experiments, we replan after each action is executed without reusing the knowledge of previous time-steps. This simple strategy helps to get a general impression of PRADA's planning performance and complexity. Other strategies are easily conceivable. For instance, one might execute the entire sequence without replanning, trading off faster computation times with a potential loss in the achieved reward. In noisy environments, it might seem a better strategy to combine the reuse of previous plans with replanning. For instance, one could omit the first action of the previous plan, which has just been executed, and examine the suitability of the remaining actions in the new state. While we consider only the single best action sequence, in many planning domains it might also be beneficial to marginalize over all sequences with the same first action. For instance, an action $a_1$ might lead to a number of reasonable sequences, none of which are the best, while another action $a_2$ is the first of one very good sequence, but also many bad ones – in which case one might favor $a_1$.

## 5.4 Illustrative Example

Let us consider the small planning problem in Table 2 to illustrate the reasoning procedure of PRADA. Our domain is a noisy cubeworld represented by predicates $table(X)$, $cube(X)$, $on(X, Y)$, $inhand(X)$ and $clear(X) \equiv \forall Y. \neg on(Y, X)$ where a robot can perform two types of actions: it may either lift a cube $X$ by means of action $grab(X)$ or put the cube which is held in hand on top of another object $X$ using $puton(X)$. The start state $s_0$ shown in 2(a) contains three cubes $a$, $b$ and $c$ stacked in a pile on table $t$. The goal shown in 2(b) is to get the middle cube $b$ on-top of the top cube $a$. Our world model provides three abstract NID rules to predict action effects, shown in Table 2(c). Only the first rule has uncertain outcomes: it models to grab an object which is below another object. In contrast, grabbing a clear object (Rule 2) and putting an object somewhere (Rule 3) always leads to the same successor state.

First, PRADA constructs a DBN to represent the planning problem. For this purpose, it computes the grounded rules with respect to the objects $\mathcal{O} = \{a, b, c, t\}$ shown in 2(d). Most potential grounded rules can be ignored: one can deduce from the abstract rules which predicates are changeable. In combination with the specifications in $s_0$, this prunes most grounded rules. For instance, we know from $s_0$ that $t$ is the table. Thus, no ground rule with action argument $X = t$ needs to be constructed as all rules require $cube(X)$.

Based on the DBN, PRADA samples action-sequences and evaluates their expected rewards. In the following, we investigate this procedure for the sampling of action-sequence $(grab(b), puton(a))$. Table 2(e) presents the inferred values of the DBN variables and other auxiliary quantities. The marginals $\alpha$ (Eq. (22)) of the state variables at $t = 0$ are set deterministically according to $s_0$. We calculate the posteriors over context variables $P(\Phi \mid \mathbf{a^{0:t-1}})$ according to Eq. (30). In our example, at $t = 0$ there is one rule with probability 1.0 for each of the actions $grab(a)$, $grab(b)$ and $grab(c)$. In contrast, there are no rules with non-zero probability for the various $puton(\cdot)$ actions. By the help of Eq. (33), we calculate the probability of each rule $r$ to be the unique covering rule for the respective action (listed under *Unique rule*; note that we do not condition on a fixed action $a^t$ thus far): this is the case if context $\Phi_r$ of $r$ holds, while all contexts $\Phi_{r'}$ of the competing rules $r'$ for the same action do not hold. At $t = 0$, this is the same as the posterior of $\Phi_r$ alone. The resulting probabilities are used to calculate the sampling distribution of Eq. (36): first, we compute the probability for each action to have a unique covering rule which is a simple sum over probabilities of the previous step (listed under *Action coverage* in the table); then, we normalize these values to get a sampling distribution $P_{sample}(\cdot)$. At $t = 0$, this results in a sampling distribution which is uniform over the three actions with unique rules. Assume we sample $a^0 = grab(b)$ (grabbing blue cube $b$). Variable $R$ specifies the ground rules to use for predicting the state marginals at the next time-step. We can infer its posterior according to Eq. (28). Here, $P(R^0 = (1, b/act) \mid a^0) = 1.0$.

Things get more interesting at $t = 1$. Here, we observe the effects of the factored frontier. For instance, consider calculating the posterior over context $\Phi_r$ for ground rule $r = (1, b/att)$ (grabbing blue cube $b$ which is below yellow $a$) using Eq. (30),

$$P(\Phi_{(1,b/att)} \mid a^0) \approx \alpha(on(a,b)) \cdot \alpha(on(b,t)) \cdot \alpha(cube(a)) \cdot \alpha(cube(b)) \cdot \alpha(table(t))$$
$$= 0.2 \cdot 0.2 \cdot 1.0 \cdot 1.0 \cdot 1.0 = 0.04.$$

In contrast, the exact value is $P(\Phi_{(1,b/att)} \mid a^0) = 0.2$, according to the third outcome of abstract Rule 1 used to predict $a^0$. The imprecision is due to ignoring the correlations: FF regards the marginals for $on(a,b)$ and $on(b,t)$ as independent, while in fact they are fully correlated.

At $t = 1$, the action $grab(a)$ has three ground rules with non-zero context probabilities (grabbing $a$ from either $b$, $c$ or $t$). This is due to the three different outcomes of abstract

## Table 2: Example of PRADA's factored frontier inference

**(a) Start state**

$s_0 = \{on(a,b), on(b,c), on(c,t),$
$cube(a), cube(b), cube(c), table(t)\}$

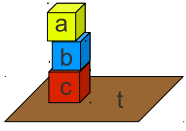

**(b) Goal**

$\tau = \{on(b,a)\}$

**(c) Abstract NID rules with example situations**

**Rule 1:**

$grab(X): \quad on(Y,X), \, on(X,Z), \, cube(X), \, cube(Y), \, table(T)$

$\rightarrow \begin{cases} 0.5: & inhand(X), \, on(Y,Z), \, \neg on(Y,X), \, \neg on(X,Z) \\ 0.3: & inhand(X), \, on(Y,T), \, \neg on(Y,X), \, \neg on(X,Z) \\ 0.2: & on(X,T), \, \neg on(X,Z) \end{cases}$

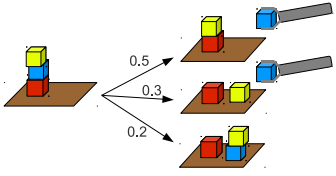

**Rule 2:**

$grab(X): \quad cube(X), \, clear(X), \, on(X,Y)$

$\rightarrow \{ \quad 1.0: \quad inhand(X), \, \neg on(X,Y)$

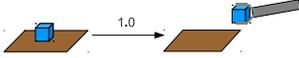

**Rule 3:**

$puton(X): \quad inhand(Y), \, cube(Y)$

$\rightarrow \{ \quad 1.0: \quad on(Y,X), \, \neg inhand(X)$

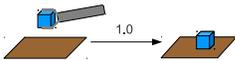

**(d) Grounded NID rules**

| Grounded Rule | Action | Substitution |
|---|---|---|
| $(1, a/bbt)$ | $grab(a)$ | $\{X \rightarrow a, Y \rightarrow b, Z \rightarrow b, T \rightarrow t\}$ |
| $(1, a/bct)$ | $grab(a)$ | $\{X \rightarrow a, Y \rightarrow b, Z \rightarrow c, T \rightarrow t\}$ |
| ... | | |
| $(1, c/bbt)$ | $grab(c)$ | $\{X \rightarrow c, Y \rightarrow b, Z \rightarrow b, T \rightarrow t\}$ |
| $(2, a/b)$ | $grab(a)$ | $\{X \rightarrow a, Y \rightarrow b\}$ |
| $(2, a/c)$ | $grab(a)$ | $\{X \rightarrow a, Y \rightarrow c\}$ |
| $(2, a/t)$ | $grab(a)$ | $\{X \rightarrow a, Y \rightarrow t\}$ |
| ... | | |
| $(2, c/t)$ | $grab(c)$ | $\{X \rightarrow c, Y \rightarrow t\}$ |
| $(3, a/b)$ | $puton(a)$ | $\{X \rightarrow a, Y \rightarrow b\}$ |
| $(3, a/c)$ | $puton(a)$ | $\{X \rightarrow a, Y \rightarrow c\}$ |
| ... | | |
| $(3, t/c)$ | $puton(t)$ | $\{X \rightarrow a, Y \rightarrow c\}$ |

**(e)** Inferred posteriors in PRADA's FF inference for action-sequence $(grab(b), puton(a))$

| | $t=0$ | $t=1$ | $t=2$ |
|---|---|---|---|
| **State marginals $\alpha$** | | | |
| $on(a,b)$ | 1.0 | 0.2 | 0.2 |
| $on(a,c)$ | 0.0 | 0.5 | 0.5 |
| $on(a,t)$ | 0.0 | 0.3 | 0.3 |
| $on(b,a)$ | 0.0 | 0.0 | 0.8 |
| $on(b,c)$ | 1.0 | 0.0 | 0.0 |
| $on(b,t)$ | 0.0 | 0.2 | 0.2 |
| $on(c,t)$ | 1.0 | 1.0 | 1.0 |
| $inhand(b)$ | 0.0 | 0.8 | 0.16 |
| $clear(a)$ | 1.0 | 1.0 | 0.2 |
| $clear(b)$ | 0.0 | 0.8 | 0.8 |
| $clear(c)$ | 0.0 | 0.5 | 0.5 |
| **Goal $U$** | 0.0 | 0.0 | 0.8 |
| **$P(\Phi \mid \mathbf{a}^{0:t-1})$** | | | |
| $\Phi_{(1, b/act)}$ | 1.0 | 0.0 | |
| $\Phi_{(1, b/att)}$ | 0.0 | 0.04 | |
| $\Phi_{(1, c/btt)}$ | 1.0 | 0.5 | |
| $\Phi_{(2, a/b)}$ | 1.0 | 0.2 | |
| $\Phi_{(2, a/c)}$ | 0.0 | 0.5 | |
| $\Phi_{(2, a/t)}$ | 0.0 | 0.3 | |
| $\Phi_{(2, b/t)}$ | 0.0 | 0.16 | |
| $\Phi_{(2, c/t)}$ | 0.0 | 0.5 | |
| $\Phi_{(3, a/b)}$ | 0.0 | 0.8 | |
| $\Phi_{(3, c/b)}$ | 0.0 | 0.8 | |
| $\Phi_{(3, t/b)}$ | 0.0 | 0.8 | |
| **Unique rule** | | | |
| $(1, b/act)$ | 1.0 | 0.0 | |
| $(1, b/att)$ | 0.0 | 0.0336 | |
| $(1, c/att)$ | 0.0 | 0.25 | |
| $(1, c/btt)$ | 1.0 | 0.0 | |
| $(2, a/b)$ | 1.0 | 0.07 | |
| $(2, a/c)$ | 0.0 | 0.28 | |
| $(2, a/t)$ | 0.0 | 0.12 | |
| $(2, b/t)$ | 0.0 | 0.154 | |
| $(2, c/t)$ | 0.0 | 0.25 | |
| $(3, a/b)$ | 0.0 | 0.8 | |
| $(3, c/b)$ | 0.0 | 0.8 | |
| $(3, t/b)$ | 0.0 | 0.8 | |
| **Action coverage** | | | |
| $grab(a)$ | 1.0 | 0.47 | |
| $grab(b)$ | **1.0** | 0.187 | |
| $grab(c)$ | 1.0 | 0.5 | |
| $puton(a)$ | 0.0 | **0.8** | |
| $puton(c)$ | 0.0 | 0.8 | |
| $puton(t)$ | 0.0 | 0.8 | |
| **Sample distribution** | | | |
| $P_{sample}(grab(a))$ | 0.33 | 0.132 | |
| $P_{sample}(grab(b))$ | **0.33** | 0.0526 | |
| $P_{sample}(grab(c))$ | 0.33 | 0.141 | |
| $P_{sample}(puton(a))$ | 0.0 | **0.225** | |
| $P_{sample}(puton(c))$ | 0.0 | 0.225 | |
| $P_{sample}(puton(t))$ | 0.0 | 0.225 | |
| **$P(R^t = r^t \mid \mathbf{a}^{0:t})$** | | | |
| $R^t = (1, b/act)$ | 1.0 | 0.0 | |
| $R^t = (3, a/b)$ | 0.0 | 0.8 | |
| $R^t = 0$ | 0.0 | 0.2 | |

Rule 1. As an example, we calculate the probability of rule $(2, a/c)$ (grabbing $a$ from $c$) to be the unique covering rule for $grab(a)$ at $t = 1$ as

$$P(\Phi_{(2,a/c)}, \neg\Phi_{(2,a/b)}, \neg\Phi_{(2,a/t)} \,|\, a^0)$$
$$\approx P(\Phi_{(2,a/c)} \,|\, a^0) \cdot (1. - P(\Phi_{(2,a/b)} \,|\, a^0)) \cdot (1. - P(\Phi_{(2,a/t)} \,|\, a^0))$$
$$= 0.5 \cdot (1. - 0.2) \cdot (1. - 0.3) = 0.28 \;.$$

After some more calculations, we determine the sampling distribution at $t = 1$. Assume we sample action $puton(a)$. This results in rule $(3/a, b)$ (putting $b$ on $a$) being used for prediction with 0.8 probability – since this is its probability to be the unique covering rule for action $puton(a)$. The remaining mass 0.2 of the posterior is assigned to those parts of the state space where no unique covering rule is available for $puton(a)$. In this case, we use the default rule $R = 0$ (corresponding to not performing the action) so that with probability 0.2 the values of the state variables persist.

Finally, let us infer the marginals at $t = 2$ using Eq. (25). As an example, we calculate $\alpha(inhand(b)^{t=2})$. Let $i(b)$ be brief for $inhand(b)$. We sum over the ground rules $r^{t=1}$ taking the potential values $i(b)^{t=1}$ and $\neg i(b)^{t=1}$ at the previous time-step $t = 1$ into account,

$$\alpha(i(b)^{t=2}) \approx \sum_{r^{t=1}} P(r^{t=1} \,|\, \mathbf{a}^{0:1}) \; ( \; P(i(b)^{t=2} \,|\, r^{t=1}, \neg i(b)^{t=1}) \; \alpha(\neg i(b)^{t=1})$$
$$+ \; P(i(b)^{t=2} \,|\, r^{t=1}, i(b)^{t=1}) \; \alpha(i(b)^{t=1}) \; )$$
$$= 0.8 \; (0.0 * 0.2 + 0.0 * 0.8) \; + \; 0.2 \; (0.0 * 0.2 + 1.0 * 0.8) = 0.16 \;.$$

As discussed above, only the ground rule $(3/a, b)$ and the default rule play a role in this prediction. In effect, the belief that $b$ is inhand decreases from 0.8 to 0.16 after having tried to put $b$ on $a$, as expected. Similarly, we calculate the posterior of $on(b, a)$ as 0.8. This is also the expected probability to reach the goal when performing the actions $grab(b)$ and $puton(a)$. (Here, PRADA's inferred value coincides with the true posterior.)

For comparison, the probability to reach the goal is 1.0 when performing the actions $grab(a)$, $puton(t)$, $grab(b)$ and $puton(a)$, i.e., when we clear $b$ before we grab it. This plan is safer, i.e., has higher probability, but takes more actions.

## 5.5 Comparison of the Planning Approaches

The most prominent difference between the presented planning approaches is in their way to account for the stochasticity of action effects. On the one hand, SST and UCT repeatedly take samples from successor state distributions and estimate the value of an action by building look-ahead trees. On the other hand, PRADA maintains beliefs over states and propagates indetermistic action effects forward. More precisely, PRADA and SST follow opposite approaches: PRADA samples actions and calculates the state transitions approximately by means of probabilistic inference, while SST considers all actions (and thus is exact in its action search) and samples state transitions. The price for considering all actions is SST's overwhelmingly large computational cost. UCT remedies this issue and samples action sequences and thus state transitions selectively: it uses previously sampled episodes to build upper confidence bounds on the estimates for action values in specific states, which are used to adapt the policy for the next episode. It is not straightforward to translate this adaptive policy to PRADA since PRADA works on beliefs over states instead of states directly. Therefore, we chose the simple policy for PRADA to sample randomly from all actions with a unique covering rule in a state (in the form of a sampling distribution to account for beliefs over states).

PRADA returns a whole plan that will transform the world state into one where the goal is fulfilled with a probability exceeding a given threshold $\theta$, in the spirit of conformant planning or probabilistic planning with no observability (Kushmerick, Hanks, & Weld, 1995). Due to their outcome-sampling, SST and UCT cannot return such a plan in a straightforward way. Instead, they provide a policy for many successor states based on their estimates of the action-values in their look-ahead tree. The estimates of states deeper in the tree are less reliable as they have been built from less episodes. If an action has been executed and a new state is observed, these estimates can be reused. Thus far, PRADA does not take any knowledge gained in previous action-sequence samples into account to adapt its policy. An elegant way to achieve this and to better exploit goal knowledge might use backpropagation through our DBNs to plan completely by inference (Toussaint & Storkey, 2006). This is beyond the scope of this paper, as it is not clear how to do this in a principled way in the large state and action spaces of relational domains. Alternatively, PRADA could give high weight to the second action of the previous best plan. Below in Sec. 5.6, we show another simple way to make use of previous episodes to find better plans.

PRADA can afford its simple action-sampling strategy as it evaluates large numbers of action-sequences efficiently and does not have to grow look-ahead trees to account for indeterministic effects. This points at an important difference: all three algorithms are faced with search spaces of action sequences which are exponential in the horizon. To calculate the value of a *given* action sequence, however, SST and UCT still need exponential time due to their outcome sampling. In contrast, PRADA propagates the state transitions forward and thus is linear in the horizon.

Like all approximate planning algorithms, neither SST, UCT nor PRADA can be expected to perform ideally in all situations. SST and UCT sample action outcomes and hence face problems if important outcomes only have small probability. For instance, consider an agent that wants to escape a room with two locked doors. If it hits the first door which is made of wood it has a chance of 0.05 to break it and escape. The second door is made of iron and has only a chance of 0.001 to break. SST and UCT may take a very long time to detect that it is 50 times better to repeatedly hit the wooden door. In contrast, PRADA recognizes this immediately after having reasoned about each of the actions once as it takes all outcomes into account. On the other hand, in PRADA's approximate inference procedure the correlations among state variables get lost while SST and UCT preserve them as they sample complete successor states. This can impair PRADA's planning performance in situations where correlations are crucial. Consider the following simple domain with two state attributes $a$ and $b$. The agent can choose from two actions modeled by the rules

$$
\begin{aligned}
action1: \quad - \quad \rightarrow \quad & \left\{ \begin{array}{lll} 0.5 & : & a,\ b \\ 0.5 & : & \neg a,\ \neg b \end{array} \right. , \text{and} \\
action2: \quad - \quad \rightarrow \quad & \left\{ \begin{array}{lll} 0.5 & : & a,\ \neg b \\ 0.5 & : & b,\ \neg a \end{array} \right. .
\end{aligned}
$$

The goal is to make both attributes either true or false, i.e., $\phi = (a \wedge b) \vee (\neg a \wedge \neg b)$. For both actions, the resulting marginals will be $\alpha(a) = 0.5$, $\alpha(\neg a) = 0.5$, $\alpha(b) = 0.5$ and $\alpha(\neg b) = 0.5$. Due to its factored frontier, PRADA cannot distinguish between both actions although $action1$ will achieve the goal, while $action2$ will not.

PRADA's estimated probabilities of states and rewards may differ significantly from their true values. This does not harm its performance in many domains as our experiments indicate (Sec. 6). We suppose the reason for this is that while PRADA's estimated probabilities can be imprecise, they enable a correct ranking of action sequences – and in planning, we are interested in choosing the best action instead of calculating correctly its value.

A further difference between the proposed algorithms is in their way to handle the noise outcome of rules: PRADA assigns very small probability to all successor states – in the spirit of the noise outcome. In contrast, for SST and UCT it does not make sense to sample from such a distribution, as any single successor state has extremely low probability and will be inadequate to estimate state and action values. Hence, they use the described workaround to assume to stay in the same state, while discounting obtained rewards.

It is straightforward for PRADA to deal with uncertain initial states. Uncertainty of initial states is common in complex environments and may for instance be caused by partial observability or noisy sensors. This uncertainty has its natural representation in the belief state PRADA works on. In contrast, SST and UCT cannot account for uncertain initial states directly, but would have to sample from the prior distribution.

### 5.6 An Extension: Adaptive PRADA

We present a very simple extension of PRADA to increase its planning accuracy. We exploit the fact that PRADA evaluates complete sequences of actions – in contrast to SST and UCT where the actions taken at $t > 0$ depend on the sampled outcomes. *Adaptive* PRADA (A-PRADA) examines the best action sequence found by PRADA. While PRADA chooses the first action of this sequence without further reasoning, A-PRADA inspects each single action of this sequence and decides by simulation whether it can be deleted. The resulting shortened sequence may lead to an increased expected reward. This is the case if actions do not have significant effects on achieving the goal or if they decrease the success probability. If such actions are omitted, the states with high reward are reached earlier and their rewards are discounted less. For instance, consider the goal to grab a blue ball: an action sequence that grabs a red cube, puts it onto the table and only then grabs the blue ball can be improved by omitting the first two actions which are unrelated to the goal.

More precisely, A-PRADA takes PRADA's action sequence $\mathbf{a}_P$ with the highest value and investigates iteratively for each action whether it can be deleted. An action can be deleted from the plan if the resulting plan has a higher reward likelihood. This idea is formalized in Algorithm 1. The crucial calculation of this algorithm is to compute values $Q(s^0, \mathbf{a}^{0:T-1})$ as defined in Eq. (28) and restated here for convenience:

$$Q(\mathbf{s}^0, \mathbf{a}^{0:T-1}) = \sum_{t=1}^{T} \gamma^t P(U^t = 1 \,|\, \mathbf{a}^{0:t-1}, \mathbf{s}^0) \,.$$

PRADA's approximate inference procedure is particularly suitable for calculating all required $P(U^t = 1 \,|\, \mathbf{a}^{0:t-1}, \mathbf{s}^0)$. It performs this calculation in time linear in the length $T$ of

---

**Algorithm 1** Adaptive PRADA (A-PRADA)

---

**Input:** PRADA's plan $\mathbf{a}_P$
**Output:** A-PRADA's plan $\mathbf{a}_A$

1: $\mathbf{a}_A \leftarrow \mathbf{a}_P$
2: **for** $t = 0$ to $t = T - 1$ **do**
3:      **while** true **do**
4:          Let $\mathbf{a}$ be a plan of length $T$.
5:          $\mathbf{a}^{0:t-1} \leftarrow \mathbf{a}_A^{0:t-1}$      $\triangleright$ Omit $a^t$
6:          $\mathbf{a}^{t:T-2} \leftarrow \mathbf{a}_A^{t+1:T-1}$
7:          $a^{T-1} \leftarrow doNothing$
8:          **if** $Q(s^0, \mathbf{a}) > Q(s^0, \mathbf{a}_A)$ **then**
9:              $\mathbf{a}_A \leftarrow \mathbf{a}$
10:         **else**
11:             break
12:         **end if**
13:      **end while**
14: **end for**
15: return $\mathbf{a}_A$

---

the action sequence, while SST and UCT would require time exponential in $T$ because of their outcome sampling.

## 6. Evaluation

We have implemented all presented planning algorithms and the learning algorithm for NID rules in C++. Our code is available at `www.user.tu-berlin.de/lang/prada/`. We evaluate our approaches in two different scenarios. The first is an intrinsically noisy complex simulated environment where we learn NID rules from experience and use these to plan. Second, we apply our algorithms on the benchmarks of the Uncertainty Part of the International Planning Competition 2008.

### 6.1 Simulated Robot Manipulation Environment

We perform experiments in a simulated complex robot manipulation environment where a robot manipulates objects scattered on a table (Fig. 4). Before we report our results in three series of experiments on different tasks of increasing difficulty, we first describe this domain in detail. We use a 3D rigid-body dynamics simulator (ODE) that enables a realistic behavior of the objects. This simulator is available at `www.user.tu-berlin.de/lang/DWSim/`. Objects are cubes and balls of different sizes and colors. The robot can grab objects and put them on top of other objects or on the table. The actions of the robot are affected by noise. In this domain, towers of objects are not straight-lined; it is easier to put an object on top of a big cube than on top of a small cube while it is difficult to put something on top of a ball; piles of objects may topple over; objects may fall off the table in which case they become out of reach for the robot.

We represent this domain with predicates $on(X, Y)$, $inhand(X)$, $upright(X)$, $out(X)$ (if an object has fallen off the table), function $size(X)$ and unary typing predicates $cube(X)$, $ball(X)$, $table(X)$. These predicates are obtained by querying the state of the simulator and translating it according to simple hand-made guidelines, thereby sidestepping the difficult problem of converting the agent's observations into an internal representation. For instance, $on(a, b)$ holds if $a$ and $b$ exert friction forces on each other and $a$'s $z$-coordinate is greater than the one of $b$, while their $x$- and $y$-coordinates are similar. Besides these primitive concepts, we also use the derived predicate $clear(X) \equiv \forall Y. \neg on(Y, X)$. We found this predicate to enable more compact and accurate rules, which is reflected in the values of the objective function of the rule learning algorithm given in Eq. (3).

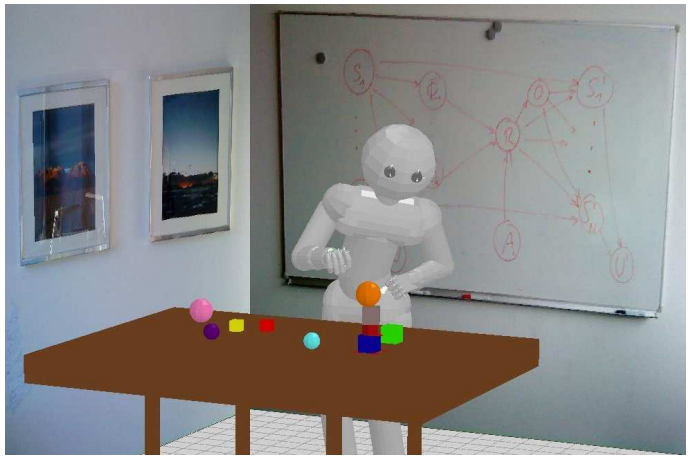

Figure 4: A simulated robot plays with cubes and balls of different sizes scattered on a table. Objects that have fallen off the table cannot be manipulated anymore.

We define three different types of actions. These actions correspond to motor primitives whose effects we want to learn and exploit. The $grab(X)$ action triggers the robot to open its hand, move its hand next to $X$, let it grab $X$ and raise the robot arm again. The execution of this action is not influenced by any further factors. For example, if a different object $Y$ has been held in the hand before, it will fall down on either the table or a third object just below $Y$; if there are objects on top of $X$, these are very likely to fall down. The $puton(X)$ action centers the robot's hand at a certain distance above $X$, opens it and raises the hand again. For instance, if there is an object $Z$ on $X$, the object $Y$ that was potentially inhand may end up on $Z$ or $Z$ might fall off $X$. The $doNothing()$ action triggers no movement of the robot's arm. The robot might choose this action if it thinks that any other action could be harmful with respect to its expected reward. We emphasize again that actions always execute, regardless of the state of the world. Also, actions which are rather unintuitive for humans such as trying to grab the table or to put an object on top of itself are carried out. The robot has to learn by itself the effects of such motor primitives.

Due to its intrinsic noise and its complexity, this simulated robot manipulation scenario is a challenging domain for both learning compact world models as well as planning. If there are $o$ objects and $f$ different object sizes, the action space contains $2o+1$ actions while the state space is huge with $f^o 2^{o^2+6o}$ different states (not excluding states one would classify as "impossible" given some intuition about real world physics).

We use the rule learning algorithm of Pasula et al. (2007) with the same parameter settings to learn three different sets of fully abstract NID rules. Each rule-set is learned from independent training sets of 500 experience triples $(s, a, s')$ that specify how the world changed from state $s$ to successor state $s'$ when an action $a$ was executed, assuming full observability. Training data to learn rules are generated in a world of six cubes and four balls of two different sizes by performing random actions with a slight bias to build high piles. Our resulting rule-sets contain 9, 10 and 10 rules respectively. These rule-sets provide approximate partial models to the true world dynamics. They generalize over the situations of the experiences, but may not account for situations that are completely different from what the agent has seen before. To enforce compactness and avoid overfitting, rules are regularized; hence, the learning algorithm may sometimes favor to model rarely experienced state transitions as low-probability outcomes in more general rules, thereby trading off accuracy for compactness. This in combination with the general noisiness of the world causes the need to carefully account for the probabilities of the world when reasoning with these rules.

We perform three series of experiments with planning tasks of increasing difficulty. In each series, we test the planners in different worlds with varying numbers of cubes and balls. Thus, we transfer the knowledge gained in the training world to different, but similar worlds by using abstract NID rules. For each object number, we create five different worlds. Per rule-set and world, we perform three independent runs with different random seeds. To evaluate the different planning approaches, we compute the mean performances and planning times over the fixed (but randomly generated) set of 45 trials (3 *learned* rule-sets, 5 worlds, 3 random seeds).

We choose the parameters of the planning algorithms as follows. For SST, we report results for different branching factors $b$, as far as the resulting runtimes allow. Similarly, UCT and (A-)PRADA each have a parameter that balances their planning time and the quality of their found actions. For UCT, this is the number of episodes, while for (A-)PRADA this is the number of sampled action-sequences. Depending on the experiment, we set both heuristically such that the tradeoff between planning time and quality is reasonable. In particular, for a fair comparison we pay attention that UCT, PRADA and A-PRADA get about the same planning times, if not reported otherwise. Furthermore, for UCT we set the bias parameter $c$ to 1.0 which we found heuristically to perform best. For all planners and experiments, we set the discounting factor for future rewards to $\gamma = 0.95$. A crucial parameter is the planning horizon $d$, which heavily influences planning time. Of course, $d$ cannot be known a-priori. Therefore, if not reported otherwise, we deliberately set $d$ larger than required for UCT and (A-)PRADA to suggest that our algorithms are also effective when $d$ can only be estimated. Indeed, we found in all our experiments that as long as $d$ is not too small, its exact choice does not have significant effects on UCT's and (A-)PRADA's planning quality – unlike its effects on planning times. In contrast, we set the horizon $d$ for SST always as small as possible, in which case its planning times are still very large. If a planning algorithm does not find a suitable action in a given situation, we restart the planning procedure: SST builds a new tree, UCT runs more episodes and (A-)PRADA takes new action-sequence samples. If in a given situation after 10 planning runs a suitable action still is not found, the trial fails.

Furthermore, we use FF-Replan (Yoon et al., 2007) as a baseline. As we discuss in more detail with the related work in Sec. 2, FF-Replan determinizes the planning problem, thereby ignoring outcome probabilities. FF-Replan has shown impressive results on the domains of the probabilistic planning competitions. These domains are carefully designed by humans: their action dynamics definitions are complete, accurate and consistent and are used as the true world dynamics in the according experiments – in contrast to the learned NID rules we use here which estimate approximate partial models of our robot manipulation domain. To be able to use the derived predicate *clear(X)* in the FF-Replan implementation of our experiments, we included the appropriate literals of this predicate by hand in the outcomes of the rules – while our SST, UCT and (A-)PRADA implementations infer these values automatically from the definition of *clear(X)*. We report results of FF-Replan with these (almost original) learned rules using the all-outcomes determinization scheme, denoted by FF-Replan-All below. (Using single-outcome schemes always led to worse performance.) Some of these rules are very general (putting only few restrictions on the arguments and deictic references); in this case, more actions appear applicable in a given state than make sense from an intuitive human perspective which hurts FF-Replan much more than the other methods, resulting in large planning times for FF-Replan. For instance, a rule may model the toppling over of a small tower including object $X$ when trying to put an object $Y$ on top of the tower: one outcome might specify $Y$ to end up below $X$. While this is only possible if $Y$ is a cube, of course, the learning algorithm may choose to omit a typing predicate *cube(X)* due to regularization, as it prefers compact rules and none of its experiences might require this additional predicate. Therefore, we created modified rule-sets by hand where we introduced typing predicates where appropriate to make contexts more distinct. Below, we denote our results with these modified rule-sets as FF-Replan-All* and FF-Replan-Single*, using all-outcomes and single most-probable outcome determinization schemes.

### 6.1.1 High Towers

In our first series of experiments, we investigate building high towers which was the planning task in the work of Pasula et al. (2007). More precisely, the reward in a state is defined as the average height of objects. This constitutes an easy planning problem as many different actions may increase the reward (object identities do not matter) and a small planning horizon $d$ is sufficient. We set SST to horizon $d=4$ (Pasula et al. 's choice) with different branching factors $b$ and UCT and (A-)PRADA to horizon $d=6$. In our experiments, initial states do not contain already stacked objects, so the reward for performing no actions is 0. Table 3 and Fig. 5 present our results. SST is not competitive. For a branching factor $b > 1$, it is slower than UCT and (A-)PRADA by at least an order of magnitude. For $b = 1$, its performance is poor. In this series of experiments, we designed the worlds of 10 objects to contain many big cubes. This explains the relatively good performance of SST in these worlds, as the number of good plans is large. As mentioned above, we control UCT, PRADA and A-PRADA to have about the same times available for planning. All three approaches perform far better than SST in almost all experiments. The difference between UCT, PRADA and A-PRADA is never significant.

This series of experiments indicates that planning approaches using full-grown look-ahead trees like SST are inappropriate even for easy planning problems. In contrast, approaches that exploit look-ahead trees in a clever way such as UCT seem to be the best choice for easy tasks which require a small planning horizon and can be solved by many alternative good plans. The performance of the planning approaches using approximate

Table 3: *High towers problem. Reward* denotes the discounted total reward for different numbers of objects (cubes/balls and table). The reward for performing no actions is 0. All data points are averages over 45 trials created from 3 learned rule-sets, 5 worlds and 3 random seeds. Standard deviations of the mean estimators are shown. *FF-Replan-All\* and FF-Replan-Single\* use hand-made modifications of the original learned rule-sets.* Fig. 5 visualizes these results.

| Objects | Planner | Reward | Trial time (s) |
|---------|---------|--------|----------------|
| 6+1 | FF-Replan-All | $6.65 \pm 1.01$ | $41.07 \pm 9.63$ |
| | FF-Replan-All* | $6.29 \pm 0.80$ | $7.54 \pm 4.09$ |
| | FF-Replan-Single* | $4.48 \pm 0.94$ | $\mathbf{4.61} \pm 2.75$ |
| | SST (b=1) | $11.68 \pm 1.19$ | $9.03 \pm 0.80$ |
| | SST (b=2) | $12.90 \pm 1.01$ | $121.40 \pm 11.12$ |
| | SST (b=3) | $12.80 \pm 0.94$ | $595.43 \pm 55.95$ |
| | UCT | $\mathbf{16.01} \pm 0.99$ | $7.45 \pm 0.19$ |
| | PRADA | $15.54 \pm 1.25$ | $6.01 \pm 0.07$ |
| | A-PRADA | $\mathbf{16.12} \pm 1.27$ | $6.36 \pm 0.07$ |
| 8+1 | FF-Replan-All | $5.10 \pm 1.01$ | $76.86 \pm 20.98$ |
| | FF-Replan-All* | $3.08 \pm 0.87$ | $28.65 \pm 16.81$ |
| | FF-Replan-Single* | $2.82 \pm 0.87$ | $\mathbf{1.72} \pm 0.27$ |
| | SST (b=1) | $9.62 \pm 1.07$ | $23.57 \pm 3.48$ |
| | SST (b=2) | $12.36 \pm 1.21$ | $335.5 \pm 52.4$ |
| | SST (b=3) | $11.09 \pm 0.87$ | $1613.3 \pm 249.2$ |
| | UCT | $\mathbf{17.11} \pm 1.07$ | $15.54 \pm 0.40$ |
| | PRADA | $16.10 \pm 1.21$ | $15.24 \pm 0.27$ |
| | A-PRADA | $16.29 \pm 1.47$ | $16.30 \pm 0.27$ |
| 10+1 | FF-Replan-All | $6.97 \pm 1.21$ | $121.99 \pm 27.43$ |
| | FF-Replan-All* | $7.36 \pm 1.07$ | $33.45 \pm 12.80$ |
| | FF-Replan-Single* | $5.76 \pm 1.21$ | $\mathbf{4.14} \pm 1.08$ |
| | SST (b=1) | $15.12 \pm 1.34$ | $119.26 \pm 10.59$ |
| | SST (b=2) | $14.48 \pm 1.20$ | $1748.7 \pm 170.2$ |
| | SST (b=3) | $16.48 \pm 1.19$ | $8424 \pm 851$ |
| | UCT | $\mathbf{17.71} \pm 1.08$ | $31.71 \pm 5.83$ |
| | PRADA | $16.21 \pm 1.07$ | $31.58 \pm 1.14$ |
| | A-PRADA | $16.78 \pm 1.14$ | $35.22 \pm 0.40$ |

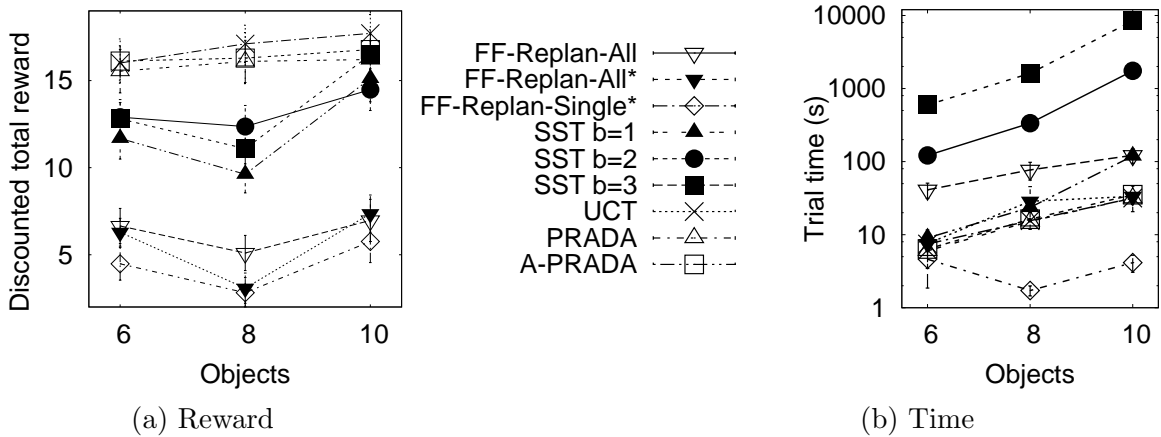

(a) Reward                                    (b) Time

Figure 5: *High towers problem* Visualization of the results presented in Table 3. The reward
for performing no actions is 0. All data points are averages over 45 trials created
from 3 learned rule-sets, 5 worlds and 3 random seeds. Error bars for the standard
deviations of the mean estimators are shown. Please note the log-scale in (b).

inference, PRADA and A-PRADA, however, comes close to the one of UCT, showing also
their suitability for such scenarios.

FF-Replan focuses on exploiting conjunctive goal structures and cannot deal with quan-
tified goals. As the grounded reward structure of this task consists of a disjunction of
different tower combinations, FF-Replan has to pick an arbitrary tower combination as its
goal. Therefore, to apply FF-Replan we sample tower combinations according to the re-
wards they achieve (i.e., situations with high towers are more probable) and do not exclude
combinations with balls at the bottom of towers as they are not prohibited by the reward
structure. As Yoon et al. note, "the obvious pitfall of this [goal formula sampling] approach
is that some groundings of the goal are not reachable or are much more expensive to reach
from the initial state". When FF-Replan cannot find a plan, we do not execute an action,
but sample a new ground goal formula at the next time-step, preserving already achieved
tower structures.

FF-Replan performs significantly worse than the previous planning approaches. The
major reason for this is that FF-Replan often comes up with plans exploiting low-probability
outcomes of rules – in contrast to SST, UCT and (A-)PRADA which reason over the
probabilities. To illustrate this, consider the example rule in Fig. 1 which models putting
a ball on top of a cube. It has two explicit outcomes: the ball usually ends up on the
cube; sometimes, however, it falls on the table. FF-Replan can misuse this rule as a tricky
way to put a ball on the table – ignoring that this often will fail. As the results of FF-
Replan-Single* show, taking only most probable outcomes into account does not remedy
this problem: there are often two to three outcomes with similar probabilities so such a
choice seems unjustified; sometimes, the "intuitively expected" outcome is split up into
different outcomes with low probabilities, which however vary only in features irrelevant for
the planning problem (such as $upright(\cdot)$).

Table 4: *Desktop clearance problem. Reward* denotes the discounted total reward for different numbers of objects (cubes/balls and table). The reward for performing no actions is 0. All data points are averages over 45 trials created from 3 learned rule-sets, 5 worlds and 3 random seeds. Standard deviations of the mean estimators are shown. *FF-Replan-All\* and FF-Replan-Single\* use hand-made modifications of the original learned rule-sets.* Fig. 6 visualizes these results.

| Obj. | Planner | Reward | Trial time (s) |
|---|---|---|---|
| | FF-Replan-All | $3.81 \pm 0.67$ | $19.1 \pm 6.5$ |
| | FF-Replan-All* | $5.86 \pm 0.87$ | $1.1 \pm 0.7$ |
| | FF-Replan-Single* | $6.53 \pm 1.07$ | $\mathbf{0.7} \pm 0.8$ |
| 6+1 | SST (b=1) | $5.35 \pm 0.75$ | $1382.6 \pm 80.4$ |
| | UCT | $9.60 \pm 0.86$ | $52.2 \pm 0.7$ |
| | PRADA | $10.94 \pm 0.86$ | $40.9 \pm 0.7$ |
| | A-PRADA | $\mathbf{12.79} \pm 0.80$ | $42.3 \pm 0.7$ |
| | FF-Replan-All | $5.93 \pm 1.00$ | $29.8 \pm 8.7$ |
| | FF-Replan-All* | $6.21 \pm 1.05$ | $3.5 \pm 0.6$ |
| | FF-Replan-Single* | $6.02 \pm 0.94$ | $\mathbf{0.8} \pm 0.7$ |
| 8+1 | SST (b=1) | $8.43 \pm 2.01$ | $8157 \pm 978$ |
| | UCT | $10.29 \pm 1.08$ | $151.4 \pm 2.0$ |
| | PRADA | $\mathbf{14.63} \pm 1.54$ | $154.5 \pm 1.9$ |
| | A-PRADA | $\mathbf{14.87} \pm 1.57$ | $157.4 \pm 2.0$ |
| | FF-Replan-All | $3.30 \pm 0.74$ | $60.9 \pm 12.1$ |
| | FF-Replan-All* | $3.53 \pm 0.87$ | $20.7 \pm 5.4$ |
| | FF-Replan-Single* | $3.91 \pm 0.86$ | $\mathbf{5.2} \pm 1.3$ |
| 10+1 | SST (b=1) | – | $> 8h$ |
| | UCT | $10.13 \pm 0.80$ | $415.7 \pm 7.4$ |
| | PRADA | $12.81 \pm 1.14$ | $385.3 \pm 4.7$ |
| | A-PRADA | $\mathbf{13.91} \pm 1.12$ | $394.5 \pm 4.0$ |

### 6.1.2 Desktop Clearance

The task in our second series of experiments is to clear up the desktop. Objects are lying splattered all over the table in the beginning. An object is cleared if it is part of a tower containing all other objects of the same class. An object class is simply defined in terms of color which is additionally provided to the state representation of the robot. The reward of the robot is defined as the number of cleared objects. In our experiments, classes contain 2-4 objects with at most 1 ball (in order to enable successful piling). Our starting situations contain some piles, but only with objects of different classes. Thus, the reward for performing no actions is 0. Desktop clearance is more difficult than building high towers, as the number of good plans yielding high rewards is significantly reduced.

We set the planning horizon $d = 6$ optimal for SST which is required to clear up a class of 4 objects, namely grabing and putting three objects. As above, by contrast we set $d = 10$ for UCT and (A-)PRADA to show that they can deal with overestimation of the usually unknown optimal horizon $d$. Table 4 and Fig. 6 present our results. The horizon $d = 6$ overburdens SST as can be seen from its large planning times. Even for $b = 1$, SST takes almost 40 minutes on average in worlds of 6 objects, while over 2 hours in worlds of 8 objects. Therefore, we did not try SST for greater $b$. In contrast, the planning times of UCT, PRADA and A-PRADA, again controlled to be about the same and to enable reasonable performance, are two orders of magnitude smaller, although overestimating the planning horizon: for a trial they take on average about $45s$ in worlds of 6 objects, $2\frac{1}{2}$ minutes in worlds of 8 objects and 6-7 minutes in worlds of 10 objects. Nonetheless, UCT, PRADA and A-PRADA perform significantly better than SST. In all worlds, PRADA and A-PRADA in turn outperform UCT, in particular in worlds with many objects. A-PRADA finds the best plans among all planners. All planners gain more reward in worlds of 8 objects in comparison to worlds of 6 objects, as the number of objects that can be cleared increases as well as the number of classes and thus of good plans. The worlds of 10 objects contain the same numbers of object classes like the worlds of 8 objects, but with more objects, making planning more difficult.

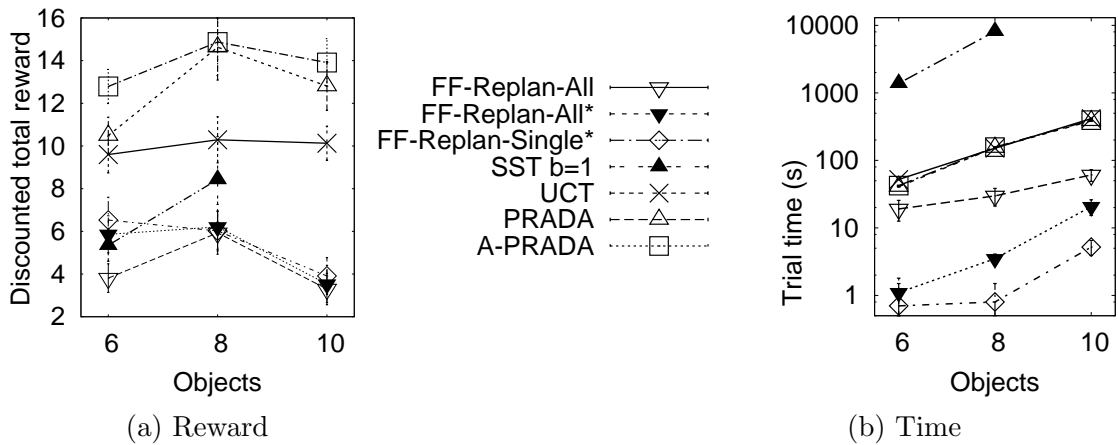

(a) Reward                               (b) Time

Figure 6: *Desktop clearance problem.* Visualization of the results presented in Table 4. The reward for performing no actions is 0. All data points are averages over 45 trials created from 3 learned rule-sets, 5 worlds and 3 random seeds. Error bars for the standard deviations of the mean estimators are shown. Note the log-scale in (b).

Overall, our findings in the *Desktop clearance* experiments indicate that while SST is inappropriate, UCT achieves good performance in planning scenarios which require medium planning horizons and where there are several, but not many alternative plans. Approaches using approximate inference like PRADA and A-PRADA, however, seem to be more appropriate in such scenarios of intermediate difficulty.

Furthermore, our results indicate that FF-Replan is inadequate for the clearance task. We sample target classes randomly to provide a goal structure to FF-Replan; the tower structure within a target class in turn is also randomly chosen. The bad performance of FF-Replan is due to the reasons described in the previous experiments; in particular the plans of FF-Replan often rely on low-probability outcomes.

Table 5: *Reverse tower problem.* The trial times and numbers of executed actions are given for the successful trials for different numbers of objects (cubes and table). All data points are averages over 45 trials created from 3 learned rule-sets, 5 worlds and 3 random seeds. Standard deviations of the mean estimators are shown. *FF-Replan-All\* and FF-Replan-Single\* use hand-made modifications of the original learned rule-sets.*

| Objects | Planner | Success rate | Trial time (s) | Executed actions |
|---------|---------|--------------|----------------|------------------|
|         | FF-Replan-All | 0.02 | 7.1 ± 0.0 | 12.0 ± 0.10 |
|         | FF-Replan-All* | **1.00** | 26.7 ± 2.7 | 13.1 ± 0.9 |
|         | FF-Replan-Single* | 0.67 | **7.0** ± 0.9 | 13.6 ± 1.1 |
| 5+1     | SST (b=1) | 0.00 | - | - |
|         | SST (b=2) | 0.00 | >1 day | - |
|         | UCT | 0.38 | 2504.9 ± 491.1 | 19.5 ± 4.0 |
|         | PRADA | 0.71 | 27.0 ± 1.8 | 13.2 ± 0.7 |
|         | A-PRADA | 0.82 | 25.4 ± 0.8 | **10.9** ± 0.8 |
|         | FF-Replan-All | 0.00 | - | - |
|         | FF-Replan-All* | **1.00** | 589.2 ± 73.7 | **12.0** ± 0.8 |
|         | FF-Replan-Single* | 0.64 | **52.7** ± 5.3 | 17.3 ± 2.1 |
| 6+1     | UCT | 0.00 | >4 h | - |
|         | PRADA | 0.47 | 66.4 ± 3.9 | 13.6 ± 0.9 |
|         | A-PRADA | 0.56 | 77.5 ± 8.3 | 14.4 ± 2.5 |
|         | FF-Replan-All | 0.00 | - | - |
|         | FF-Replan-All* | **0.42** | 2234.2 ± 81.1 | **15.1** ± 1.3 |
| 7+1     | FF-Replan-Single* | 0.56 | **687.4** ± 86.4 | 17.5 ± 2.0 |
|         | PRADA | 0.24 | 871.3 ± 126.6 | 18.2 ± 1.2 |
|         | A-PRADA | 0.23 | 783.7 ± 132.6 | **15.1** ± 1.8 |

### 6.1.3 Reverse Tower

To explore the limits of UCT, PRADA and A-PRADA, we conducted a final series of experiments where the task is to reverse towers of $C$ cubes which requires at least $2C$ actions (each cube needs to be grabbed and put somewhere at least once). Apart from the long planning horizon, this is difficult due to the noise in the simulated world: towers can become unstable and topple over with cubes falling off the table. To decrease this noise slightly to obtain more reliable results, we forbid the robot to grab objects that are not clear (i.e., below other objects). We set a limit of 50 executed actions on each trial. If thereafter the reversed tower still is not built, the trial fails. The trial also fails if one of the required objects falls off the table.

Table 5 presents our results. We cannot get SST with optimal planning horizon $d = 10$ to solve this problem even for five cubes. Although the space of possible actions is reduced due to the mentioned restriction, SST has enormous runtimes. With $b = 1$, SST does not find suitable actions (no leaves with the goal state) in several starting situations – the increased planning horizon leads to a high probability of sampling at least one unfavorable outcome for a required action. For $b \geq 2$, a single tree traversal of SST takes more than a day. We found UCT to also require large planning times in order to achieve a reasonable success rate. Therefore, we set the planning horizons optimal for UCT. In worlds of 5 cubes, UCT with optimal $d = 10$ has a success rate of about 40% while taking on average more than 40 minutes in case of success. For 6 cubes, however, UCT with optimal $d = 12$ never succeeds even when planning times exceed 4 hours. In contrast, we can afford an overestimating horizon $d = 20$ for PRADA and A-PRADA. In worlds of 5 cubes, PRADA and A-PRADA achieve success rates of 71% and 82% respectively in less than half a minute. A-PRADA's average number of executed actions in case of success is almost optimal. In worlds of 6 cubes, the success rates of PRADA and A-PRADA are still about 50%, taking a bit more than a minute on average in case of success. When their trials fail, this is most often due to cubes falling off the table and not because they cannot find appropriate actions. Cubes falling off the table is also a main reason why the success rates of PRADA and A-PRADA drop to 23% and 24% respectively in worlds of 7 cubes when towers become rather unstable. Planning times in successful trials, however, also increase to more than 13 minutes indicating the limitations of these planning approaches. Nonetheless, the mean number of executed actions in successful trials is still almost optimal for A-PRADA.

Overall, the *Reverse tower* experiments indicate that planning approaches using look-ahead trees fail in tasks that require long planning horizons and can only be achieved by very few plans. Given the huge action and state spaces in relational domains, the chances that UCT simulates an episode with exactly the required actions and successor states are very small. Planning approaches using approximate inference like PRADA and A-PRADA have the crucial advantage that the stochasticity of actions does not affect their runtime exponentially in the planning horizon. Of course, their search space of action-sequences still is exponential in the planning horizon so that problems requiring long horizons are hard to solve also for them. Our experiments show that by using the very simple, though principled extension A-PRADA, we can gain significant performance improvements.

Our results also show that FF-Replan fails to provide good plans when using the original learned rule-sets. This is surprising as the characteristics of the *Reverse tower* task seem to favor FF-Replan in comparison to the other methods: there is a single conjunctive goal structure and the number of good plans is very small while these plans require long horizons. As the results of FF-Replan-All* and FF-Replan-Single* indicate, FF-Replan can achieve a good performance with the adapted rule-sets that have been modified by hand to restrict the number of possible actions in a state. While this constitutes a proof of concept of FF-Replan, it shows the difficulty of applying FF-Replan with learned rule-sets.

### 6.1.4 SUMMARY

Our results demonstrate that successful planning with learned world models (here in the form of rules) may require to explicitly account for the quantification of predictive uncertainty. More concretely, methods applying look-ahead trees (UCT) and approximate inference ((A-)PRADA) outperform FF-Replan on different tasks of varying difficulty. Furthermore, (A-)PRADA can solve planning tasks with long horizons, where UCT fails. Only if one post-processes the learned rules by hand to clarify their application contexts and the planning problem uses a conjunctive goal structure and requires few and long plans, FF-Replan performs better than UCT and (A-)PRADA.

## 6.2 IPPC 2008 Benchmarks

In the second part of our evaluation, we apply our proposed approaches on the benchmarks of the latest international probabilistic planning competition, the Uncertainty Part of the International Planning Competition in 2008 (IPPC, 2008). The involved domains differ in many characteristics, such as the number of actions, the required planning horizons and the reward structures. As the competition results show, no planning algorithm performs best everywhere. Thus, these benchmarks give an idea for what types of problems SST, UCT and (A-)PRADA may be useful. We convert the PPDDL domain specifications into NID rules along the lines described in Sec. B.1. The resulting rule-sets are used to run our implementations of SST, UCT and (A-)PRADA on the benchmark problems.

Each of the seven benchmark domains consists of 15 problem instances. An instance specifies a goal and a starting state. Instances vary not only in problem size, but also in their reward structures (including action costs), so a direct comparison is not always possible. In the competition, each instance was considered independently: planners were given a restricted amount of time (10 minutes for problems 1-5 of each domain and 40 minutes for the others) to cover as many repetitions of the very same problem instance as possible up to a maximum of a 100 trials. Trials differed in the random seeds resulting in potentially different state transitions. The planners were evaluated with respect to the number of trials ending in a goal state and the collected reward averaged over all trials.

Eight planners entered in the competition, including FF-Replan which was not an official participant. They are discussed with the related work in Sec. 2. For their results, which are too voluminous to be presented here, we refer the reader to the website of the competition. Below, we provide a *qualitative* comparison of our methods to the results of these planners. We do not attempt a direct *quantitative* comparison for several reasons. First, the different hardware prevents timing comparisons. Second, competition participants have frequently not been able to successfully cover trials of a single or all instances of a domain. It is difficult to tell the reasons for this from the results tables: the planner might have been overburdened by the problem, might have faced temporary technical problems with the client-server architecture framework of the competition or could not cope with certain PPDDL constructs which could have been rewritten in a simpler format.

Third and most importantly, we have not optimized our implementations to reuse previous planning efforts. Instead, we fully replan for each single action (within a trial and across trials). The competition evaluation scheme puts replanners at a disadvantage (in particular those which replan each single action). Instead of replanning, a good strategy for the competition is to spend most planning time before starting the first trial and then reuse the resulting insights (such as conditional plans and value functions) for all subsequent trials with a minimum of additional planning. Indeed, this strategy has often been adopted as many trial time results indicate. We acknowledge that this is a fair procedure to evaluate planners which compute policies over large parts of the state-space before acting. We feel, however, that this is counter to the idea of our approaches: UCT and (A-)PRADA are meant for flexible planning with varying goals and different situations. Thus, what we are interested in is the average time to compute good actions and successfully solve a problem instance when there is no prior knowledge available.

Table 6: *Benchmarks of the IPPC 2008.* The first column of a table specifies the problem instance. *Suc.* is the success rate. The trial time and the number of executed actions are given for the successful trials. Where applicable, the reward for all trials is shown. All results are achieved with *full replanning* within a trial and across trials.

(a) Search and Rescue

| | Planner | Suc. | Trial Time (s) | Actions | Reward |
|---|---|---|---|---|---|
| 01 | SST | 100 | 37.9±0.1 | 9.2±0.2 | 1440±90 |
| | UCT | 54 | 1.4±0.1 | 11.4±0.3 | 900±70 |
| | PRADA | 100 | 1.1±0.1 | 10.5±0.4 | 1460±89 |
| | A-PRADA | 100 | 1.1±0.1 | 10.4±0.4 | 1460±89 |
| 02 | SST | 100 | 220.2±0.1 | 9.8±0.2 | 1560±83 |
| | UCT | 56 | 4.1±0.3 | 12.2±0.6 | 880±100 |
| | PRADA | 100 | 1.6±0.1 | 12.9±0.7 | 1460±89 |
| | A-PRADA | 100 | 1.6±0.1 | 12.8±0.4 | 1440±90 |
| 03 | SST | 71 | 955.5±0.5 | 9.8±0.2 | 1662±85 |
| | UCT | 57 | 12.9±0.6 | 13.6±0.6 | 680±63 |
| | PRADA | 99 | 1.4±0.1 | 18.0±1.0 | 1480±88 |
| | A-PRADA | 99 | 1.4±0.1 | 17.9±1.1 | 1480±88 |
| 04 | UCT | 61 | 24.9±1.6 | 16.1±0.8 | 7200±57 |
| | PRADA | 100 | 1.4±0.0 | 11.9±0.4 | 1460±89 |
| | A-PRADA | 100 | 1.4±0.0 | 11.5±0.3 | 1500±87 |
| 05 | UCT | 46 | 40.1±2.1 | 16.8±1.4 | 600±64 |
| | PRADA | 89 | 6.8±0.3 | 21.8±0.9 | 1240±83 |
| | A-PRADA | 92 | 6.5±0.3 | 21.0±0.9 | 1320±81 |
| 06 | UCT | 39 | 71.7±5.6 | 19.5±1.3 | 410±59 |
| | PRADA | 83 | 10.1±0.9 | 24.3±1.3 | 1240±90 |
| | A-PRADA | 84 | 10.0±0.9 | 23.7±1.2 | 1240±90 |
| 07 | UCT | 53 | 230.3±13.2 | 21.5±1.4 | 540±62 |
| | PRADA | 98 | 10.1±0.4 | 18.5±0.8 | 1470±88 |
| | A-PRADA | 98 | 9.9±0.4 | 18.0±0.8 | 1490±87 |
| 08 | UCT | 34 | 332.9±24.1 | 21.71±1.5 | 360±59 |
| | PRADA | 59 | 20.2±0.8 | 30.4±1.7 | 910±82 |
| | A-PRADA | 59 | 19.9±0.8 | 29.9±1.7 | 910±82 |
| 09 | UCT | 30 | 752.8±72.3 | 26.4±2.4 | 360±48 |
| | PRADA | 63 | 30.2±1.2 | 27.5±1.6 | 930±80 |
| | A-PRADA | 65 | 30.0±1.1 | 27.5±1.6 | 1010±84 |
| 10 | PRADA | 21 | 97.9±10.2 | 26.8±2.8 | 180±27 |
| | A-PRADA | 21 | 92.1±9.8 | 26.7±2.8 | 180±27 |
| 11 | PRADA | 17 | 151.7±12.3 | 30±2.5 | 250±29 |
| | A-PRADA | 18 | 154.1±11.9 | 30.2±2.6 | 250±29 |
| 12 | PRADA | 38 | 210.8±72.1 | 30.1±10.5 | 636±253 |
| | A-PRADA | 21 | 219.8±28.5 | 30.7±2.8 | 556±55 |

(b) Triangle-Tireworld

| | Planner | Suc. | Trial Time (s) | Actions |
|---|---|---|---|---|
| 01 | SST | 0 | – | – |
| | UCT | 100 | 9.9±0.3 | 6.9±0.2 |
| | PRADA | 100 | 8.5±0.2 | 6.4±0.2 |
| | A-PRADA | 100 | 8.0±0.2 | 6.1±0.2 |
| 02 | UCT | 100 | 64.1±2.2 | 12.4±0.3 |
| | PRADA | 57 | 30.1±0.7 | 9±0.2 |
| | A-PRADA | 65 | 33.7±0.8 | 11.4±0.3 |
| 03 | UCT | 89 | 390.5±8.5 | 18.6±0.4 |
| | PRADA | 19 | 119.2±4.9 | 12.3±0.5 |
| | A-PRADA | 21 | 121.0±5.3 | 14.3±0.7 |
| 04 | UCT | 82 | 1497±19 | 26.0±0.5 |
| | PRADA | 6 | 2967±143 | 17.5±1.1 |
| | A-PRADA | 4 | 244.2±43.6 | 15.5±2.8 |

(c) Blocksworld

| | Planner | Suc. | Trial Time (s) | Actions | Reward |
|---|---|---|---|---|---|
| 01 | SST | 0 | – | – | – |
| | UCT | 0 | – | – | – |
| | PRADA | 53 | 17.8±0.4 | 23.0±0.7 | 0.8±0.0 |
| | A-PRADA | 63 | 18.4±0.5 | 22.3±0.8 | 0.6±0.0 |
| 03 | PRADA | 10 | 57.0±3.3 | 21.5±1.8 | -9.6±0.0 |

(d) Boxworld

| | Planner | Suc. | Trial Time (s) | Actions | Reward |
|---|---|---|---|---|---|
| 01 | SST | 0 | – | – | – |
| | UCT | 0 | – | – | – |
| | PRADA | 100 | 257.8±6.3 | 46.8±1.0 | 1.00±0.0 |
| | A-PRADA | 100 | 143.8±3.1 | 43.1±1.1 | 1.00±0.0 |
| 02 | PRADA | 100 | 285.2±7.8 | 46.2±1.3 | 20.00±0.0 |
| | A-PRADA | 100 | 215.8±4.2 | 39.6±0.9 | 20.00±0.0 |
| 03 | UCT | 100 | 1285.2±8.1 | 32.8±0.0 | 929.8±2.1 |
| | PRADA | 100 | 165.7±2.9 | 52.5±1.1 | 865.1±3.3 |
| | A-PRADA | 50 | 457.8±7.1 | 35.0±0.7 | 754.1±21.5 |
| 04 | PRADA | 28 | 959.0±35.5 | 76.1±3.2 | 0.3±0.5 |
| | A-PRADA | 60 | 519.2±15.3 | 72.0±2.4 | 0.6±0.1 |
| 05 | UCT | 54 | 9972±776 | 37.9±3.5 | 606±149 |
| | PRADA | 61 | 345.4±8.5 | 68.4±1.6 | 465±24 |
| | A-PRADA | 2 | 528.6±38.8 | 38.0±0.0 | 411±34 |
| 08 | PRADA | 3 | 3361±88 | 87.0±2.3 | 0.19±0.1 |
| | A-PRADA | 10 | 1579±48 | 85.3±2.7 | 0.29±0.3 |
| 09 | PRADA | 28 | 1449±25 | 85.9±1.5 | 1365±31 |
| | A-PRADA | 0 | – (1750.3) | – | 1126±30 |

(e) Exploding Blocksworld

| | Planner | Suc. | Trial Time (s) | Actions |
|---|---|---|---|---|
| 01 | SST | 5 | 8607±1224 | 9.6±0.6 |
| | UCT | 3 | 111.8±14.0 | 9.3±0.4 |
| | PRADA | 62 | 3.6±0.0 | 8.6±0.8 |
| | A-PRADA | 61 | 3.9±0.0 | 8.4±0.8 |
| 02 | PRADA | 28 | 11.9±0.3 | 14.4±0.5 |
| | A-PRADA | 29 | 12.7±0.2 | 13.2±0.5 |
| 03 | PRADA | 36 | 14.3±0.3 | 12.6±0.6 |
| | A-PRADA | 30 | 16.8±0.3 | 12.5±0.5 |
| 04 | PRADA | 27 | 30.3±1.2 | 14.8±0.5 |
| | A-PRADA | 26 | 14.9±1.1 | 15.2±0.5 |
| 05 | PRADA | 100 | 5.5±0.1 | 6.6±0.1 |
| | A-PRADA | 100 | 5.5±0.1 | 6.6±0.1 |
| 06 | PRADA | 51 | 128.5±2.9 | 16.9±0.7 |
| | A-PRADA | 61 | 97.5±5.3 | 17.3±0.8 |
| 07 | PRADA | 14 | 125.0±6.9 | 15.3±0.4 |
| | A-PRADA | 72 | 154.8±5.5 | 17.6±1.0 |

Therefore, for each single problem instance we perform 100 trials with different random seeds using full replanning. A trial is aborted if a goal state is not reached within some maximum number of actions varying slightly for each benchmark (about 50 actions). We present the success rates and the mean estimators of trial times, executed actions and rewards with their standard deviations in Table 6 for the problem instances where at least one trial was successfully covered in reasonable time.

**Search and Rescue** (Table 6(a)) is the only domain where SST (with branching factor 1) is able to find plans within reasonable time – with significantly larger runtimes than UCT and (A-)PRADA. The success rates and the rewards indicate that PRADA and A-PRADA are superior to UCT and scale up to rather big problem instances. To give an idea w.r.t. the IPPC evaluation scheme: UCT solves successfully 54 trials of the first instance within 10 minutes with *full replanning*, while PRADA and A-PRADA solve all trials with full replanning. In fact, despite of replanning each single action, PRADA and A-PRADA show the same success rates as the best planners of the benchmark except for the very large problem instances (within the competition, only the participants FSP-RBH and FSP-RDH achieved comparably satisfactory results). We conjecture that the success of our methods is due to that fact that this domain requires to account carefully for the outcome probabilities, but does not involve very long planning horizons.

**Triangle-Tireworld** (Table 6(b)) is the only domain where UCT outperforms PRADA and A-PRADA, although at a higher computational cost. The more depth-first-like style of planning of UCT seems useful in this domain. To give an idea w.r.t. the IPPC evaluation scheme: UCT performs 60 successful trials of the first instance within 10 minutes, while PRADA and A-PRADA achieve 72 and 74 trials resp. using full replanning; but UCT solves more trials in the more difficult instances. The required planning horizons increase quickly with the problem instances. Our approaches cannot cope with the large problem instances, which only three competition participants (RFF-BG, RFF-PG, HMDPP) could cover.

Our methods face problems when the required planning horizons are very large, while the number of plans with non-zero probability is small. This becomes evident in the **Blocksworld** benchmark (Table 6(c)). This domain is different from the robot manipulation environment of our first evaluation in Sec. 6.1. The latter is considerably more stochastic and provides more actions in a given situation (e.g., we may grab objects within a pile). Blocksworld is the only domain where our approaches are inferior to FF-Replan. To give an idea w.r.t. the IPPC evaluation scheme: UCT does not perform a single successful trial of the first instance within 10 minutes, while PRADA and A-PRADA achieve 16 and 17 trials resp. using full replanning.

In the **Boxworld** domain (Table 6(d)), our approaches can exploit the fact that the delivery of boxes is (almost) independent of the delivery of other boxes (in most problem instances this is further helped by the intermediate rewards for delivered boxes). In contrast to UCT, PRADA and A-PRADA scale up to relatively large problem instances. PRADA and A-PRADA solve all 100 trials of the first problem instance, requiring on average 4.3 min and 2.4 min resp. with full replanning. Only two competition participants solved trials successfully in this domain (RFF-BG and RFF-PG). To give an idea w.r.t. the IPPC evaluation scheme: UCT does not perform a single successful trial within 10 minutes, while PRADA completes 2 and A-PRADA 4 trials. This small number can be explained by the large plan lengths where each single action is computed with full replanning.

Finally, in the **Exploding Blocksworld** domain (Table 6(e)) PRADA and A-PRADA perform better or as good as the competition participants. To give an idea w.r.t. the IPPC evaluation scheme: UCT achieves only a single successful trial within 10 minutes, while PRADA and A-PRADA complete 56 and 61 trials resp..

We did not perform any experiments in either the **SysAdmin** or the **Schedule** domain. Their PPDDL specifications cannot be converted into NID rules due to the involved universal effects. In contrast, this has been possible for the Boxworld domain despite of the universal effects there: in the Boxworld problem instances, the universally quantified variables always refer to exactly one object which we exploit for conversion to NID rules. (Note that this can be understood as a trick to implement deictic references in PPDDL by means of universal effects. The according action operator, however, has odd semantics: boxes could end up in two different cities at the same time.) Furthermore, we ignored the **Rectangle-Tireworld** domain, which together with the Triangle-Tireworld domain makes up the 2-Tireworlds benchmark, as its problem instances have faulty goal descriptions: They should include *not(dead)* (this has not been critical to name a winner in the competition as personally communicated by Olivier Buffet).

### 6.2.1 Summary

The majority of the PPDDL descriptions of the IPPC benchmarks can be converted into NID rules, indicating the broad spectrum of planning problems which can be covered by NID rules. Our results demonstrate that our approaches perform comparably to or better than state-of-the-art planners on many traditional hand-crafted planning problems. This hints at the generality of our methods for probabilistic planning beyond the type of robotic manipulation domains considered in Sec. 6.1. Our methods perform particularly well in domains where outcome probabilities need to be carefully accounted for. They face problems when the required planning horizons are very large, while the number of plans with non-zero probability is small; this can be avoided by intermediate rewards.

## 7. Discussion

We have presented two approaches for planning with probabilistic relational rules in grounded domains. Our methods are designed to work on *learned* rules which provide approximate partial models of noisy worlds. Our first approach is an adaptation of the UCT algorithm which samples look-ahead trees to cope with action stochasticity. Our second approach, called PRADA, models the uncertainty over states explicitly in terms of beliefs and employs approximate inference in graphical models for planning. When we combine our planning algorithms with an existing rule learning algorithm, an intelligent agent can *(i)* learn a compact model of the dynamics of a complex noisy environment and *(ii)* quickly derive appropriate actions for varying goals. Results in a complex simulated robotics domain show that our methods outperform the state-of-the-art planner FF-Replan on a number of different planning tasks. In contrast to FF-Replan, our methods reason over the probabilities of action outcomes. This is necessary if the world dynamics are noisy and only partial and approximate world models are available.

However, our planners also perform remarkably well on many traditional probabilistic planning problems. This is demonstrated by our results on IPPC benchmarks, where we have shown that PPDDL descriptions can be converted to a large extent to the kind of rules our planners use. This hints at the general-purpose character of particularly PRADA and the potential benefits of its techniques for probabilistic planning. For instance, our methods can be expected to perform similarly well in large propositional MDPs which do not exhibit a relational structure.

So far, our planning approaches deal in reasonable time with problems containing up to 10-15 objects (implying billions of world states) and requiring planning horizons of up to 15-20 time-steps. Nonetheless, our approaches are still limited in that they rely on reasoning in the grounded representation. If very many objects need to be represented or if the representation language gets very rich, our approaches need to be combined with other methods that reduce state and action space complexity (Lang & Toussaint, 2009b).

## 7.1 Outlook

In its current form, the approximate inference procedure of PRADA relies on the specific compact DBNs compiled from rules. The development of similar factored frontier filters for arbitrary DBNs, e.g. derived from more general PPDDL descriptions, is promising. Similarly, the adaptation of PRADA's factored frontier techniques into existing probabilistic planners is worth of investigation.

Using probabilistic relational rules for backward planning appears appealing. It is straightforward to learn NID rules that regress actions by providing reversed triples $(s', a, s)$ to the rule learning algorithm, stating the predecessor state $s$ for a state $s'$ if an action $a$ has been applied before. Backward planning, which can be combined with forward planning, has received a lot of attention in classical planning and may be fruitful for both planning with look-ahead trees as well as planning using approximate inference. By means of propagating backwards through our DBNs, one may ultimately derive algorithms that calculate posteriors over actions, leading to true planning by inference (instead of sampling actions).

An important direction for improving our PRADA algorithm is to make it adapt its action-sequence sampling strategy to the experience of previous samples. We have introduced a very simple extension, A-PRADA, to achieve this, but more sophisticated methods are conceivable. Learning rule-sets online and exploiting them immediately by our planning method is also an important direction of future research in order to enable acting in the real world, where we want to behave effectively right from the start. Improving the rule framework for more efficient and effective planning is another interesting issue. For instance, instead of using a noisy default rule, one may use mixture models to deal with actions with several (non-unique) covering rules, or in general use parallel rules that work on different hierarchical levels or different aspects of the underlying system.

## Acknowledgments

We thank the anonymous reviewers for their careful and thorough comments which have greatly improved this paper. We thank Sungwook Yoon for providing us an implementation of FF-Replan. We thank Olivier Buffet for answering our questions on the probabilistic planning competition 2008. This work was supported by the German Research Foundation (DFG), Emmy Noether fellowship TO 409/1-3.

## Appendix A. Proof of Proposition 1

**Proposition 1 (Sec. 5.3)** *The set of action sequences PRADA samples with non-zero probability is a super-set of the ones of SST and UCT.*

**Proof:** Let $\mathbf{a}^{0:T-1}$ be an action sequence that was sampled by SST (or UCT). Thus, there exists a state sequence $\mathbf{s}^{0:T}$ and a rule sequence $\mathbf{r}^{0:T-1}$ such that in every state $s^t$ ($t < T$), action $a^t$ has a unique covering rule $r^t$ that predicts the successor state $s^{t+1}$ with probability $p^t > 0$. For, if $p^t = 0$, then $s^{t+1}$ would never be sampled by SST (or UCT).

We have to show that $\forall t, 0 \leq t < T : P(s^t \,|\, \mathbf{a}^{0:t-1}, s^0) > 0$. If this is the case then $P_{sample}^t(a^t) > 0$ as $a^t$ has the unique covering rule $r^t$ in $s^t$ and $a^t$ will eventually be sampled. $P(s^0) = 1 > 0$ is obvious. Now assume $P(s^t \,|\, \mathbf{a}^{0:t-1}, s^0) > 0$. If we execute $a^t$, we will get $P(s^{t+1} \,|\, \mathbf{a}^{0:t}, s^0) \geq p^t P(s^t \,|\, \mathbf{a}^{0:t-1}, s^0) > 0$. The posterior $P(s^{t+1} \,|\, \mathbf{a}^{0:t}, s^0)$ can be greater (first inequality) due to persistence or to previous states having non-zero probability that also lead to $s^{t+1}$ given $a^t$.

The set of action sequences PRADA samples is larger than that of SST (or UCT) as SST (or UCT) refuses to model the noise outcomes of rules. Assume an action $a$ and state $s$ to be the only state where $a$ has a unique covering rule. If an episode to $s$ can only be simulated by means of rule predictions with the noise outcome, this action will never be sampled by SST (or UCT) (as the required states are never sampled). In contrast, PRADA also models the effects of the noise outcome by giving very low probability to all possible successor states with the heuristic described above. $\square$

## Appendix B. Relation between NID rules and PPDDL

We use NID rules (Sec. 3.2) as relational model of the transition dynamics of probabilistic actions. Besides allowing for negative literals in the preconditions, NID rules extend probabilistic STRIPS operators (Kushmerick et al., 1995; Blum & Langford, 1999) by two special constructs, namely deictic references and noise outcomes, which are crucial for learning compact rule-sets. An alternative language to specify probabilistic relational planning problems used by the International Probabilistic Planning Competitions (IPPC, 2008) is the *probabilistic planning domain definition language* (PPDDL) (Younes & Littman, 2004). PPDDL is a probabilistic extension of a subset of PDDL, derived from the deterministic *action description language* (ADL). ADL, in turn, introduced universal and conditional effects and negative precondition literals into the (deterministic) STRIPS representation. Thus, PPDDL allows for the usage of syntactic constructs which are beyond the expressive power of NID rules; however, many PPDDL descriptions can be converted into NID rules.

Before taking a closer look at how to convert PPDDL and NID rule representations into each other, we clarify what is meant by "action" in each of the formalisms, giving an intuition of the line of thinking when using either of these. We understand by "abstract action" an abstract action predicate, e.g. $pickup(X)$. Intuitively, this defines a certain type of action. The stochastic state transitions according to an abstract action can be specified by both abstract NID rules as well as abstract PPDDL action operators (also called schemata). Typically, several different abstract NID rules model the same abstract action, specifying state transitions in different contexts. In contrast, usually only one abstract PPDDL action operator is used to model an abstract action: context-dependent effects are modeled by means of conditional and universal effects.

To make predictions in a specific situation for a concrete action (a grounded action predicate such as $pickup(greenCube)$), the strategy within the NID rule framework is to ground the set of abstract NID rules and examine which ground rules cover this state-action pair. If there is exactly one such ground rule, it is chosen for prediction. If there is no such rule or if there is more than one (the contexts of NID rules do not have to be mutually exclusive), one chooses the noisy default rule, essentially saying that one does not know what will happen (other strategies are conceivable, but not pursued here). In contrast, as there is usually exactly one operator per abstract action in PPDDL domains, there is no need of the concept of operator uniqueness and to distinguish between ground actions and operators.

## B.1 Converting PPDDL to NID rules

In the following, we discuss how to convert PPDDL features into a NID rule representation. While it may be impossible to convert a PPDDL action operator into a single NID rule, one may often translate it into a *set* of rules with at most a polynomial increase in the size of representation. Table 7 provides an example of a converted PPDDL action operator of the IPPC domain *Exploding Blocksworld*. As NID rules support many, but not all of the features a sophisticated domain description language such as PPDDL provides, using rules will not lead to compact representations in all possible domains. Our experiments, however, show that the dynamics of many interesting planning domains can be specified compactly. Furthermore, additional expressive power in rule contexts can be gained by using derived predicates which allow to bring in various kinds of logical formulas such as quantification.

**Conditional Effects**  A conditional effect in a PPDDL operator takes the form *when C then E*. It can be accounted for by two NID rules: the first rule adds $C$ to its context and $E$ to its outcomes, while the second adds $\neg C$ to its context and ignores $E$.

**Universal Effects**  PPDDL allows to define universal effects. These specify effects for all objects that meet some preconditions. An example is the *reboot* action of the *SysAdmin* domain of the IPPC 2008 competition: it specifies that every computer other than the one rebooted can independently go down with probability 0.2 if it is connected to a computer that is already down. This *cannot* be expressed in a NID rule framework. While we can refer to objects other than the action arguments via deictic references, we require these deictic references to be unique. For the *reboot* action, we would need a unique way to refer to each other computer which cannot be achieved without significant modifications (for example, such as enumerating the other computers via separate predicates).

**Disjunctive Preconditions and Quantification**  PPDDL operators allow for disjunctive preconditions, including implications. For instance, the *Search-and-rescue* domain of the IPPC 2008 competition defines an action operator $goto(X)$ with the precondition $(X \neq base) \rightarrow humanAlive()$. A disjunction $A \lor B$ ($\equiv \neg A \rightarrow B$) can be accounted for by either using two NID rules, with the first rule having $A$ in the context and the second rule having $\neg A \land B$. Alternatively, one may introduce a derived predicate $C \equiv A \lor B$. In general, the "trick" of derived predicates allows to overcome syntactical limitations of NID

Table 7: Example for converting a PPDDL action operator into NID rules. The *putDown*-operator of the IPPC benchmark domain *Exploding Blocksworld* (a) contains a conditional effect which can be accounted for by two NID rules which either exclude (b) or include (c) this condition in their context.

(a)

( : *action putDown*

  : *parameters* (?b − block)

  : *precondition* (and (holding ?b) (noDestroyedTable))

  : *effect* (and (emptyhand) (onTable ?b) (not (holding ?b))

    (probabilistic 2/5 (when (noDetonated ?b) (and (not (noDestroyedTable)) (not (noDetonated?b))))))

)

(b)

$putDown(X):$   $block(X),\ holding(X),\ noDestroyedTable(),\ \neg noDetonated(X)$

    $\rightarrow$ { $1.0$   :   $emptyhand(X),\ onTable(X),\ \neg holding(X)$

(c)

$putDown(X):$   $block(X),\ holding(X),\ noDestroyedTable(),\ noDetonated(X)$

    $\rightarrow$ $\begin{cases} 0.6 & : & emptyhand(X),\ onTable(X),\ \neg holding(X) \\ 0.4 & : & emptyhand(X),\ onTable(X),\ \neg holding(X),\ \neg noDestroyedTable(),\ \neg noDetonated(X) \end{cases}$

rules and bring in various kinds of logical formulas such as quantifications. As discussed by Pasula et al. (2007), derived predicates are an important prerequisite to being able to learn compact and accurate rules.

**Types**   Terms may be typed in PPDDL, e.g. $driveTo(C − city)$. Typing of objects and variables in predicates and functions can be achieved in NID rules by the usage of typing predicates within the context, e.g. using an additional predicate $city(C)$.

**State Transition Rewards**   In PPDDL, one can encode Markovian rewards associated with state transitions (including action costs as negative rewards) using fluents and update rules in action effects. One can achieve this in NID rules by associating rewards with the outcomes of rules.

## B.2 Converting NID rules to PPDDL

We show in the following that the way NID rules are used in SST, UCT and PRADA at *planning* time can be handled via at most a polynomial blowup in representational size. The basic building blocks of a NID rule, i.e. the context as well as the outcomes, transfer one-to-one to PPDDL action operators. The deictic references, the uniqueness requirement of covering rules and the noise outcome need special attention.

**Deictic References**   Deictic references in NID rules allow to refer to objects which are not action arguments. In PPDDL, one can refer to such objects by means of universal conditional effects. There is an important restriction, however: a deictic reference needs to pick out a single unique object in order to apply. If it picks out none or many, the rule fails to apply. There are two ways to ensure this uniqueness requirement within PPDDL. First, if allowing quantified preconditions, an explicit uniqueness precondition for each deictic reference $D$ can be introduced. Using universal quantification, it constrains all objects satisfying the preconditions $\Phi_D$ of $D$ to be identical, i.e., $\forall X, Y : \Phi_D(X, *) \wedge \Phi_D(Y, *) \rightarrow X = Y$, where $*$ are some other variables. Alternatively, uniqueness of deictic references can be achieved by a careful planning problem specification, which however cannot be guaranteed when learning rules.

**Uniqueness of covering rules** The contexts of NID rules do not have to be mutually exclusive. When we want to use a rule for prediction (as in planning), we need to ensure that it uniquely covers the given state-action pair. The procedural evaluation process for NID rules can be encoded declaratively in PPDDL using modified conditions which explicitly negate the contexts of competing rules. For instance, if there are three NID rules with potentially overlapping contexts A, B, and C (propositional for simplicity), the PPDDL action operator may define four conditions: $c_1 = \{A \wedge \neg B \wedge \neg C\}$, $c_2 = \{\neg A \wedge B \wedge \neg C\}$, $c_3 = \{\neg A \wedge \neg B \wedge C\}$, $c_4 = \{(\neg A \wedge \neg B \wedge \neg C) \vee (A \wedge B) \vee (A \wedge C) \vee (B \wedge C)\}$. Conditions $c_1$, $c_2$ and $c_3$ test for uniqueness of the corresponding NID rules and subsume their outcomes. Condition $c_4$ tests for non-uniqueness (either no covering rule or multiple covering rules) and models potential changes as noise, analogous to the situations in a NID rule context in which the noisy default rule would be used.

**Noise outcome** The noise outcome of a NID rule subsumes seldom or utterly complex outcomes. It relaxes the frame assumption: even not explicitly stated things may change with a certain probability. This comes at the price of the difficulty to ensure a well-defined successor state distribution $P(s' \mid s, a)$. In contrast, PPDDL needs to explicitly specify everything that might change. This may be an important reason why it is difficult to come up with an effective learning algorithm for PPDDL.

While in principle PPDDL does not provide for a noise outcome, the way our approaches account for it in *planning* can be encoded in PPDDL. We either treat the noise outcome as having no effects (in SST and UCT; basically a noop operator then) which is trivially translated to PPDDL; or we consider the probability of each state attribute to change independently (in PRADA) which can be encoded in PPDDL with independent universal probabilistic effects.

The noise outcome allows to always make predictions for an arbitrary action: if there are no or multiple covering rules, we may use the (albeit not very informative) prediction of the default rule. Such cases can be dealt with in PPDDL action operators using explicit conditions as described in the previous paragraph.